\documentclass[sigconf]{acmart}

\usepackage[super]{nth}
\usepackage{amsmath}
\usepackage{bm}
\usepackage{array}
\usepackage{makecell}
\usepackage{multirow}
\usepackage{tabularx}
\usepackage{graphicx}
\usepackage{caption}
\usepackage{subcaption}
\usepackage{url}
\usepackage{float}
\usepackage[flushleft]{threeparttable}
\usepackage{enumitem}

\AtBeginDocument{%
  \providecommand\BibTeX{{%
    \normalfont B\kern-0.5em{\scshape i\kern-0.25em b}\kern-0.8em\TeX}}}

\copyrightyear{2021} 
\acmYear{2021} 
\setcopyright{acmcopyright}\acmConference[CIKM '21]{Proceedings of the 30th ACM International Conference on Information and Knowledge Management}{November 1--5, 2021}{Virtual Event, QLD, Australia}
\acmBooktitle{Proceedings of the 30th ACM International Conference on Information and Knowledge Management (CIKM '21), November 1--5, 2021, Virtual Event, QLD, Australia}
\acmPrice{15.00}
\acmDOI{10.1145/3459637.3482487}
\acmISBN{978-1-4503-8446-9/21/11}

\begin{document}
\title{Tree Decomposed Graph Neural Network}

\author{Yu Wang}
\email{yu.wang.1@vanderbilt.edu}
\affiliation{%
  \institution{Vanderbilt University}
  \country{}
}

\author{Tyler Derr}
\email{tyler.derr@vanderbilt.edu}
\affiliation{%
  \institution{Vanderbilt University}
  \country{}
}



\begin{abstract}
Graph Neural Networks (GNNs) have achieved significant success in learning better representations by performing feature propagation and transformation iteratively to leverage neighborhood information. Nevertheless, iterative propagation restricts the information of higher-layer neighborhoods to be transported through and fused with the lower-layer neighborhoods', which unavoidably results in feature smoothing between neighborhoods in different layers and can thus compromise the performance, especially on heterophily networks. Furthermore, most deep GNNs only recognize the importance of higher-layer neighborhoods while yet to fully explore the importance of multi-hop dependency within the context of different layer neighborhoods in learning better representations. In this work, we first theoretically analyze the feature smoothing between neighborhoods in different layers and empirically demonstrate the variance of the homophily level across neighborhoods at different layers. Motivated by these analyses, we further propose a tree decomposition method to disentangle neighborhoods in different layers to alleviate feature smoothing among these layers. Moreover, we characterize the multi-hop dependency via graph diffusion within our tree decomposition formulation to construct Tree Decomposed Graph Neural Network (TDGNN), which can flexibly incorporate information from large receptive fields and aggregate this information utilizing the multi-hop dependency. Comprehensive experiments demonstrate the superior performance of TDGNN on both homophily and heterophily networks under a variety of node classification settings. Extensive parameter analysis highlights the ability of TDGNN to prevent over-smoothing and incorporate features from shallow layers with deeper multi-hop dependencies, which provides new insights towards deeper graph neural networks. The implementation of TDGNN is available at \url{https://github.com/YuWVandy/TDGNN}. 
\end{abstract}

\keywords{
graph neural networks, 
tree decomposition, multi-hop dependency  \vspace{-3ex}}


\maketitle

\section{Introduction}\label{sec-introduction}
Graph representation learning has recently emerged as a powerful strategy for node classification~\cite{hamilton2020graph, GAT, GCN, EGCN}, graph classification~\cite{hamilton2020graph, DGI, GIN, Diffpool} and link prediction~\cite{multilink, SEAL} on graph-structured data. As the generalization of deep learning to the graph domain, Graph Neural Networks (GNNs) have become one of the most promising paradigms~\cite{rong2020deep}, which adopts a neighborhood aggregation scheme to learn node representations by utilizing both the node features and the graph topology~\cite{GAT, SGC, GCN}. A typical GNN architecture for node classification consists of two stages: propagation/aggregation and transformation. First, messages are propagated from neighboring nodes to their corresponding center nodes and then aggregated together. Afterwards, the aggregated messages are transformed by the transformation layer to extract useful node representations. These two stages are packed together and termed as one layer of graph convolution. Deep GNNs iteratively perform multiple graph convolutions to obtain a larger receptive field and thus incorporate information of neighborhoods in higher layers~\cite{DAGNN, SGCC, GCNII, GRAND}. 

Although GNNs have gained significant achievements, a common challenge faced by GNNs is known as the over-smoothing problem~\cite{Delioversmoothing, Oversmoothing}: the performance of GNNs degrades when stacking multiple graph convolution layers. Most popular models, such as GCN~\cite{GCN} and GAT~\cite{GAT}, achieve their best performance with 2-layer graph convolutions. 
Such shallow architectures limit their ability to extract information from higher-layer neighborhoods. However, stacking multiple layers to increase the receptive field tends to fuse representations of nodes from different classes and thus make them indistinguishable due to iterative propagation~\cite{DAGNN}. Earlier works have found that the stationary point that node representations converge to is determined by node degrees and their features~\cite{Oversmoothing, proxmity}. 
However, these works focus on the theoretical analysis of the steady state in the limit of propagation while yet to provide effective solutions to solve the over-smoothing problem.

Stepping further, several methods propose deep GNNs to incorporate higher-layer neighborhood information through iterative propagation~\cite{SGC, SGCC, APPNP, Diffusion}.
 For example, GCNII~\cite{GCNII} applies an initial residual connection and identity mapping to enable GCN to express a $K^\text{th}$-order polynomial filter with arbitrary coefficients.  
Nevertheless, more recently, it has been analyzed that the performance degradation due to over-smoothing is because the entanglement of feature transformation and propagation~\cite{DAGNN}. 
Building upon this, DAGNN~\cite{DAGNN} proposes to decouple the feature propagation and transformation, and learn node representations by adaptively incorporating information from larger receptive fields. 
However, the propagation in all of previous deep models is executed iteratively so that the information of higher-layer neighborhoods is restricted to be transported through and fused with the lower-layer neighborhoods' and then propagated to their corresponding center nodes, which unavoidably informs feature smoothing between neighborhoods in different layers.

\indent Therefore, although different deep GNNs~\cite{DAGNN, GCNII, GRAND, GPR} develop their own techniques, their core component is to apply multiple graph convolutions (mixed-order propagation~\cite{GRAND}) to incorporate more neighborhood information from a broader neighboring range. As a result, the importance of depth of GNNs has raised significant concern while the importance of width of GNNs is rarely researched. It is shown in~\cite{GRAND} that the mixed order propagation rule enables their model to incorporate more local information but does not clarify the relationship between the propagation rule and the local information. Additionally, in~\cite{wang2020direct}, they observe that nodes can not only attend to their immediate neighbors but can also extract useful information from their multi-hop neighboring context. However, this work considers this multi-hop neighboring context in a specific attention framework. Different from previous works, we propose the general concept of multi-hop dependency as a measurement of the width of GNNs, and empirically demonstrate its importance in learning node representations.

\indent In view of the challenge that features of higher-layer neighborhoods are over-smoothed with the lower-layer neighborhoods and noticing the importance of the multi-hop dependency, we propose an effective framework, termed as Tree Decomposed Graph Neural Network (TDGNN), to learn node representations from larger receptive fields without causing feature over-smoothing between different layers of neighborhoods and allow flexible layer configurations to avoid under-performance on heterophily networks. 
Our major contributions are listed as follows:
\begin{itemize}
    \item Motivated by our theoretical analysis on feature smoothing and empirically demonstration of the variance of the homophily level across neighborhoods in different layers, we propose a tree decomposition method to disentangle features of neighborhoods in different layers, which can help alleviate the problem of feature smoothing and provides more flexible layer configurations for complex networks. 
    
    \item We capture and maintain the importance of multi-hop dependency in learning better representations within our tree decomposition method by characterizing this multi-hop dependency by graph diffusion, which ultimately leads to the construction of proposed Tree Decomposed Graph Neural Network (TDGNN). 
    
    \item We conduct experiments in both semi-supervised and full-supervised settings and on both homophily and heterophily network datasets to comprehensively demonstrate the superiority of our proposed TDGNN framework over existing methods. Additionally, we perform a parameter analysis to better understand and contrast TDGNN to prior GNNs.
\end{itemize}

\indent The rest of the paper is organized as follows. In Section \ref{sec-preliminaries}, we define necessary notations and briefly introduce the supervised node classification problem and GNNs. We present our proposed TDGNN framework in Section~\ref{sec-framework}, which consists of the tree decomposition procedure to disentangle feature information of neighborhoods in different layers, the graph diffusion procedure to model multi-hop dependencies, and layer aggregation procedure to enable adaptive combination of aggregated representations of different layers. In Section~\ref{sec-experiments}, experiments are performed to evaluate the effectiveness of our framework. Related work is then presented in Section~\ref{sec-relatedwork}. Finally, we conclude and discuss future work in Section \ref{sec-conclusion}.

\section{Preliminaries}\label{sec-preliminaries}
In this section, we first introduce the notations and definitions that are used throughout this paper, and then provide a brief background on the supervised node classification problem and GNNs.

\subsection{Notations}
Let $\mathcal{G} = (\mathcal{V}, \mathcal{E}, \mathbf{X})$ denote an unweighted and undirected network, where $\mathcal{V} = \{v_1, v_2, ..., v_n\}$ is the set of $n$ nodes (i.e., $n = |\mathcal{V}|$), $\mathcal{E} \subset \mathcal{V} \times \mathcal{V}$ is the set of $m$ edges (i.e., $m = |\mathcal{E}|$) between nodes in $\mathcal{V}$ with $e_{ij}$ denoting the edge between the node $v_i$ and node $v_j$, and $\mathbf{X}\in\mathbb{R}^{n\times d}$ denotes the node feature matrix, where each row $\mathbf{x}_i \in \mathbb{R}^d$ represents the feature vector of node $v_i$ and $d$ is the dimension of node features. 
The topological information of the whole network $\mathcal{G}$ is described by the adjacency matrix $\mathbf{A}\in \mathbb{R}^{n\times n}$, where $\mathbf{A}_{ij} = 1$ if an edge exists between node $v_i$ and node $v_j$ (i.e., if $e_{ij} \in \mathcal{E}$), and $\mathbf{A}_{ij} = 0$ otherwise. The diagonal matrix of node degrees are notated as $\mathbf{D}\in\mathbb{R}^{n\times n}$, where the degree of the node $v_i$ is calculated by $\mathbf{D}_{ii} = \sum_{j}{\mathbf{A}_{ij}}$. Additionally, we let $\widetilde{\mathbf{A}} = \mathbf{A} + \mathbf{I}$ represent the adjacency matrix with added self-loops and similarly let $\widetilde{\mathbf{D}}$ represent the diagonal degree matrix with the diagonal element $\widetilde{\mathbf{D}} = \mathbf{D} + \mathbf{I}$. $\mathcal{N}_i$ is the neighborhood node set of the center node $v_i$, which is given by $\mathcal{N}_i = \{v_j|e_{ij}\in\mathcal{E}\}$. We then extend this definition by using $\mathcal{N}_i^l$ to denote the $l$-th layer neighborhood nodes of the center node $v_i$, which includes all nodes that can be reached from the center node $v_i$ in exactly $l$ hops.

\subsection{Supervised Node Classification Task} 
In this work, we focus on the node classification task and leave the application of our framework on other tasks, such as link prediction, as one future work. More specifically, we consider the transductive node classification problem where we are provided with the labeled node set $\mathcal{V}_l\subset \mathcal{V}$ associated with the node label matrix $\mathbf{Y}\in\mathbb{R}^{|\mathcal{V}_l|\times C}$ with $C$ number of classes, the goal is to learn a mapping $F$: $\mathbb{R}^{n\times d}\times \mathbb{R}^{n\times n} \rightarrow \mathbb{R}^{n\times C}$, which takes as input the feature matrix $\mathbf{X}$ and adjacency matrix $\mathbf{A}$, and outputs the predicted $C$-dimensional node representation $\mathbf{Z}\in\mathbb{R}^{n\times C}$.

\begin{figure*}[t]
\centering
\includegraphics[width=.98\textwidth]{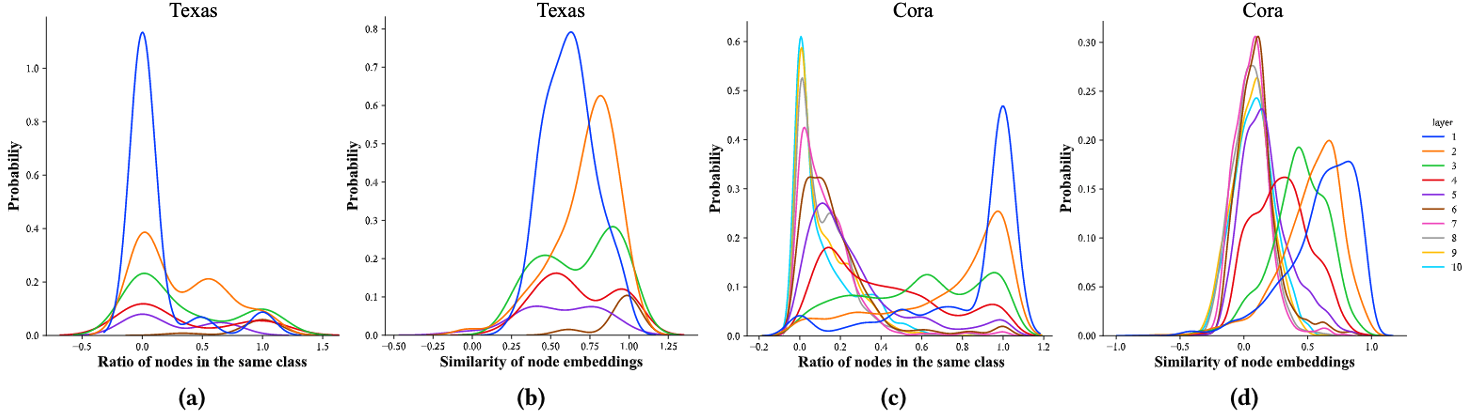}
\vskip -1.75ex
\caption{Visualizing the variance of homophily across neighborhoods at different levels according to the distribution of the ratio of different layer neighborhoods in the same class as their corresponding center nodes (i.e., (a) and (c)) and the cosine similarity of their embeddings (obtained from feeding node features through only the transformation layers of a pre-trained 2-layer GCN) to their center nodes (i.e., (b) and (d)) 
for the Texas and Cora datasets.}
     \label{fig-distcora}
\vspace{-1.75ex}
\end{figure*}

\subsection{Graph Neural Networks}
Typically, most graph neural networks (GNNs) can be decomposed into two operational procedures: (1) neighborhood propagation and aggregation, and (2) feature transformation. The neighborhood propagation and aggregation can be formalized as follows:
\begin{equation}\label{eq-propagation}
    \widehat{\mathbf{h}}_i^l = \text{AGGREGATION}^{l}(\mathbf{h}_{i}^{l - 1}, \{\mathbf{h}_j^{l - 1}|j\in\mathcal{N}_i\}),
\end{equation}
where representations of neighborhoods at the previous layer $\{\mathbf{h}_j^{l - 1}|j\in\mathcal{N}_i\}$ are propagated to the center node $v_i$ and further fused with its own representation $\mathbf{h}_i^{l - 1}$ from the previous layer via $\text{AGGREGATION}^l$ function at layer $l$ to get the partial representation $\widehat{\mathbf{h}}_i^l$. Note that $\mathbf{h}_i^0$ of node $v_i$ is initialized as the original node feature $\mathbf{x}_i$. Then, after the aggregation procedure, the $\text{TRANSFORMATION}^l$ function at layer $l$ is applied on $\widehat{\mathbf{h}}_i^l$ to get the representation $\mathbf{h}_i^l$ of node $v_i$ at layer $l$ and defined as follows:
\begin{equation}\label{eq-transformation}
    \mathbf{h}_i^l = \text{TRANSFORMATION}^{l}(\widehat{\mathbf{h}}_i^l),
\end{equation}
where we can denote the node representations at $l^{\text{th}}$ layer for all the nodes in the network as $\mathbf{H}^l\in\mathbb{R}^{n\times d^{l}}$, where the $i^{\text{th}}$ row corresponds to the representation of the node $v_i$ at layer $l$ (i.e., $\mathbf{h}_i^l \in \mathbb{R}^{d^{l}}$).

 Most graph convolutions, such as GCN \cite{GCN}, GraphSAGE \cite{Graphsage}, GAT \cite{GAT}, GIN \cite{GIN}, and SGC \cite{SGC}, can be obtained under this framework by adopting and configuring different functions in AGGREGATION and TRANSFORMATION. For instance, the vanilla GCN model suggests using $\widehat{\mathbf{H}}^l = \widehat{\mathbf{A}}\mathbf{H}^{l - 1}$ in $\text{AGGREGATION}$ followed by $\mathbf{H}^{l} = \sigma(\widehat{\mathbf{H}}^l\mathbf{W}^{l})$ in TRANSFORMATION, where $\widehat{\mathbf{A}} = \widetilde{\mathbf{D}}^{-\frac{1}{2}}\widetilde{\mathbf{A}}\widetilde{\mathbf{D}}^{-\frac{1}{2}}$ is the renormalized adjacency matrix to prevent gradient explosion, $\mathbf{W}^l\in\mathbb{R}^{d^{l - 1}\times d^{l}}$ represents the weight matrix at layer $l$ transforming features of dimension $d^{l-1}$ to $d^{l}$, and $\sigma$ is ReLU. Although GNNs may differ in their unique AGGREGATION and TRANSFORMATION designs, most graph convolutions employ an iterative propagation to increase their receptive field and incorporate information of neighborhoods in higher layers. However, such iterative propagation inevitably causes feature smoothing between neighborhoods in different layers. Furthermore, the enhanced performance caused by leveraging iterative feature propagation guides us to pay more attention to the importance of higher-layer neighborhoods while the effect of multi-hop dependency still remains unclear.
 
Next, having defined the basic notations, background, and discussed some of the challenges for supervised node classification and basic GNNs, we present our proposed framework with the purpose of alleviating the feature smoothing between neighborhoods in different layers and incorporating the multi-hop dependency.

\section{The proposed framework}\label{sec-framework}
\begin{figure*}
     \centering
     \includegraphics[width=.98\textwidth]{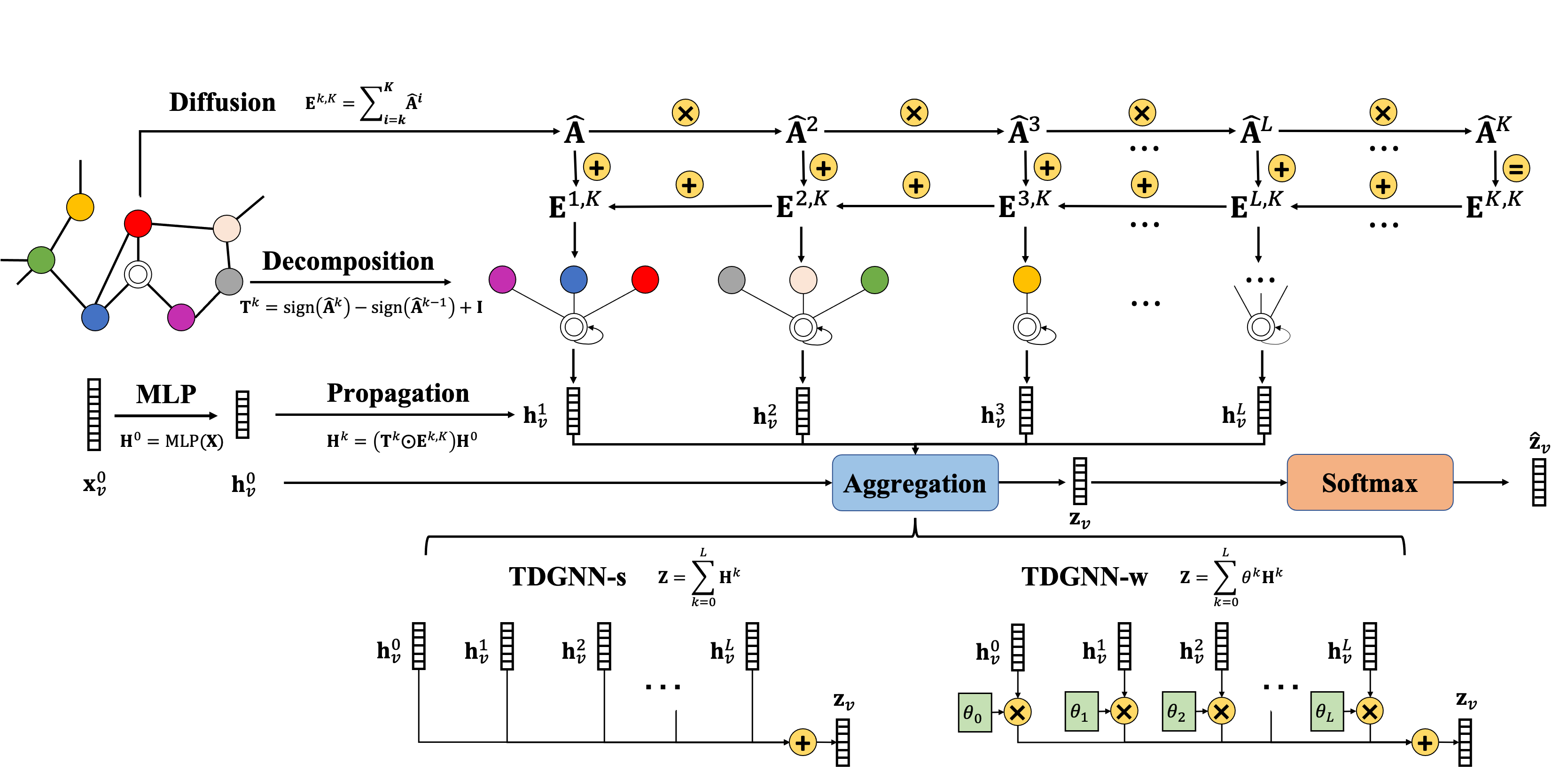}
     \vskip -1.5ex
     \caption{An illustration of the proposed Tree Decomposed Graph Neural Network (TDGNN). For brevity, the pipeline to generate the prediction for only one node is presented.}
     \label{fig-framework}
     \vspace{-1.5ex}
\end{figure*}

In this section, we design a Tree Decomposed Graph Neural Network (TDGNN) by mainly solving the two challenges mentioned in Section~\ref{sec-introduction}, which are feature smoothing between neighborhoods in different layers and lack of considering the multi-hop dependency in GNNs. For the first challenge, we theoretically show the feature smoothing between different layers when applying iterative propagation and further propose a tree decomposition method to disentangle neighborhood information in different layers. For the second challenge, we formalize the definition of the multi-hop dependency and characterize it through a graph diffusion process. Combining the tree decomposition method to disentangle the neighborhood information on different layers and the graph diffusion to model the multi-hop dependency, we propose TDGNN. We also introduce two mechanisms to aggregate node representations of each layer, TDGNN-s that directly sums the representations of all layers together, and TDGNN-w that assigns learnable weights and adaptively combines the node representations of each layer. 
The whole framework is shown in Figure~\ref{fig-framework}, which has three main components: tree decomposition to handle feature smoothing between different neighborhood layers, graph diffusion to model multi-hop dependency, and aggregation to combine representations of different layers. Next, we describe each of these components in detail.

\vspace{-1ex}
\subsection{Tree Decomposition}\label{sec-treedecomposition}
The basic assumption in GNNs is that the neighborhood information of the center node leveraged by applying feature propagation and aggregation can enhance the prediction performance of the center node itself~\cite{JK}. Such an assumption is justified by the core network property, homophily, where linked nodes tend to share similar features and typically belong to the same class~\cite{zhu2020graph, geomgcn}. However, the level of homophily might be completely different among different networks or even vary among different subgraphs within the same network. One extreme situation would be the heterophily network where linked nodes are likely from different classes or have dissimilar features~\cite{pandit2007netprobe, zhu2020beyond}.

In Figure~\ref{fig-distcora}, we show the level of homophily across different neighborhood layers in the Cora and Texas datasets. More specifically, the level of homophily is measured by distributions of the ratio of neighborhoods $\mathcal{N}^l$ in different neighborhood layers $l$ that share the same class as (in Figure~\ref{fig-distcora}(a) and Figure~\ref{fig-distcora}(c)) and have similar embeddings to (in Figure~\ref{fig-distcora}(b) and Figure~\ref{fig-distcora}(d)) their corresponding center nodes, which are obtained from feeding node features only through the transformation layers in a pre-trained 2-layer GCN~\cite{GCN} without introducing any bias from feature propagation. In the Cora dataset, it can be observed that the majority of neighbors among their $1^\text{st}$-layer neighborhoods have the same class as their corresponding center nodes, but the number of center nodes that have most of their neighborhoods sharing the same class as themselves decreases as the layer increases. Furthermore, the embeddings of these neighborhoods on the $1^\text{st}$-layer, on average, have high similarity to their corresponding center nodes. This demonstrates the high homophily of the Cora dataset on low layers and propagating features of nodes in these low layers fuses embeddings of nodes in the same class and thus makes embeddings of different classes more separable. 
However, even for this extreme homophily layer in this strong homophily Cora dataset~\cite{coin}, not all of the nodes in the $1^\text{st}$-layer have all of their neighborhoods sharing the same label with themselves and this strong homophily becomes progressively weaker as we reach further out to higher layers. For example, only around 10\% of the nodes in the $3^{\text{rd}}$-layer have all their neighborhoods sharing the same class and over half of the nodes in the $10^{\text{th}}$-layer have nearly all neighborhoods different from themselves. Even worse, in Texas dataset, even for low layers, most of the neighborhoods have different classes from their center nodes, especially for nodes on the $1^\text{st}$-layer, almost all nodes belong to different classes from their center nodes, which demonstrates the strong heterophily of the Texas dataset. Propagating features of nodes in such low layers to their center nodes fuses embeddings of nodes in different classes and makes those nodes indistinguishable, which results in learning worse node representations. Such feature smoothing among different layers is unavoidable as long as the procedure of iterative propagation is taken, that is: \textit{during iterative feature propagation, the information of neighborhoods in higher layers has to be transported through and fused with the information of neighborhoods in lower layers and then propagated to their corresponding center nodes}. 

If we take the most popular GNN-variant,
a 2-layer GCN, as an example (for simplicitly), then after 2-layer graph convolutions, the representation of the node $v_i$ is:
\begin{align}\label{eq-layerfusion}
    \mathbf{h}_i^2 &= \underbrace{\sigma(\mathbf{h}_i^{0}\mathbf{W}^{0})\mathbf{W}^{1}(\frac{1}{(d_i + 1)^2} + \sum_{j\in\mathcal{N}_i}{\frac{1}{(d_i + 1)(d_j + 1)}})}_{\text{$0^\text{th}$-layer features (self)}} \nonumber
    \\& + \underbrace{\sum_{j\in\mathcal{N}_i}{\sigma(\mathbf{h}_j^0\mathbf{W}^{0})\mathbf{W}^{1}}(\frac{d_i + d_j +2}{(d_i + 1)^{1.5}(d_j + 1)^{1.5}})}_{\text{$1^\text{st}$-layer neighborhood features}} \nonumber
    \\& + \underbrace{\sum_{j\in\mathcal{N}_i}{\sum_{k\in\{\mathcal{N}_j\cap\mathcal{N}_i\}}{\sigma(\mathbf{h}_k^0\mathbf{W}^{0})\mathbf{W}^{1}}\frac{1}{\sqrt{d_i + 1}\sqrt{d_k + 1}(d_j + 1)}}}_{\text{$1^\text{st}$-layer neighborhood features}}\nonumber
    \\& + \underbrace{\sum_{j\in\mathcal{N}_i}{\sum_{k\in \{ \mathcal{N}_j\cap\mathcal{N}_i^2 \} }{\sigma(\mathbf{h}_k^0\mathbf{W}^{0})\mathbf{W}^{1}}\frac{1}{\sqrt{d_i + 1}\sqrt{d_k + 1}(d_j + 1)}}}_{\text{$2^\text{nd}$-layer neighborhood features}},
\end{align}
which consists of three components corresponding to the feature information of neighborhoods in the $0^\text{th}, 1^\text{st}$, and $2^\text{nd}$ layers. Eq.~\ref{eq-layerfusion} intuitively shows that the representation of node $v_i$ after two iterative graph convolutions contains the information of both $1^\text{st}$ and $2^{\text{nd}}$- layer neighborhood information, which will compromise the performance on Texas dataset since the feature and class information of the $1^\text{st}$-layer is of great difference from their corresponding center nodes according to Figure~\ref{fig-distcora}(a) and Figure~\ref{fig-distcora}(b). Even for the Cora dataset, where the feature and class information of the $1^\text{st}$-layer neighborhoods is similar to their corresponding center nodes, still some center nodes have neighborhoods in the $1^\text{st}$-layer different from themselves and for these nodes, incorporating their neighborhood information might compromise their predictions.

\begin{figure}[t]
     \centering
     \includegraphics[width=0.46\textwidth]{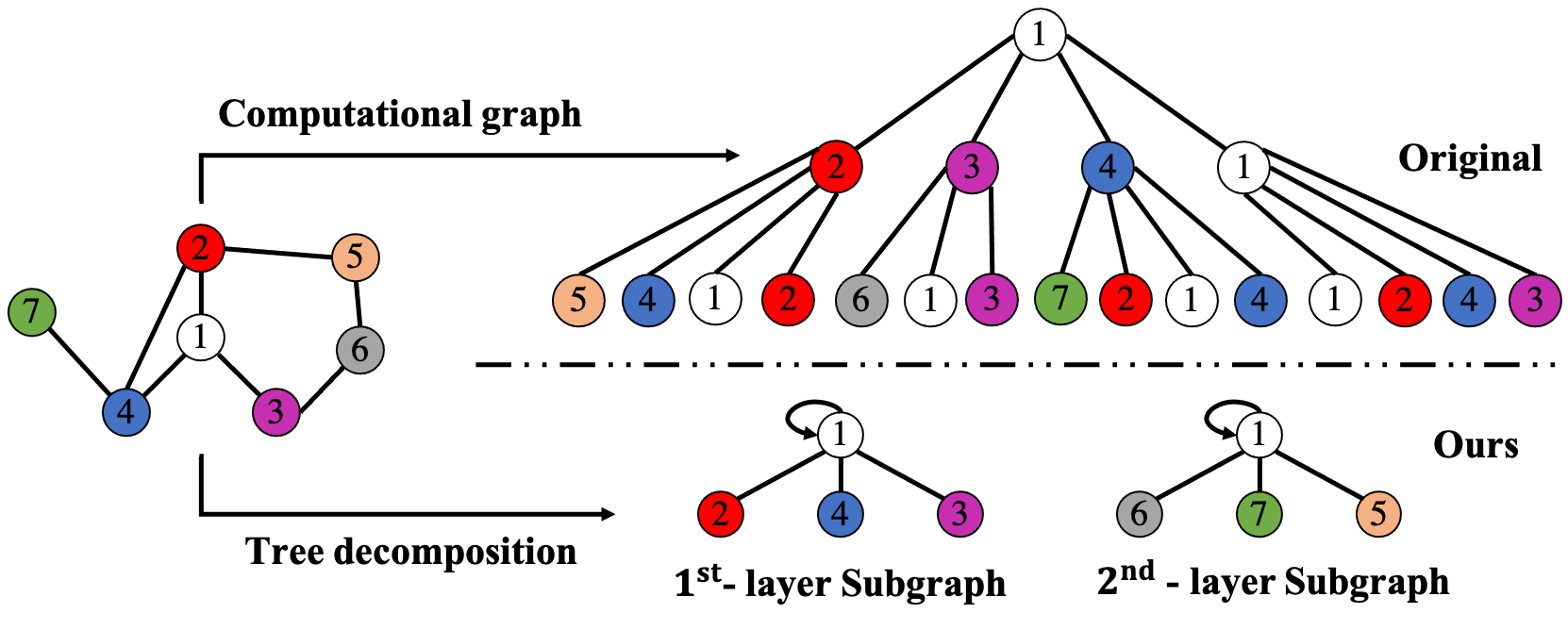}
     \vskip -1.5ex
     \caption{Tree decomposition of the center node $v_1$ in the given graph to two layers compared to the computational graph in the original GNNs (e.g., GCN).}
     \label{fig-tree-decomposition}
     \vspace{-2.5ex}
\end{figure}

Thus, based on our analysis, to advance the frontier of GNNs to be able to selectively leverage neighborhood information in different layers, we propose a tree decomposition method. More specifically, our proposed method disentangles neighborhoods in different layers and connects them directly with their corresponding center nodes. These direct connections allow the propagation of higher-layer neighborhoods' features to their corresponding center nodes without any interference of lower-layer neighborhoods along the way. Furthermore, this tree decomposition procedure enables more flexible layer configurations of neighborhoods. For example in Figure \ref{fig-tree-decomposition}, we decompose the computational tree in GNNs of the center node $v_1$. Then in the training process, we selectively propagate features of nodes in different layers: propagating along $1^\text{st}$-layer subgraph, along $2^\text{nd}$-layer subgraph, and along both of these two subgraphs. The choice depends on the network homophily of different layers and is determined by hyperparameter-tuning. The adjacency matrix of the $k^{\text{th}}$-layer subgraph $\mathbf{T}^{k}$ obtained from tree decomposition can be computed by the difference between corresponding powers of the normalized adjacency matrices with added self-loops and formalized as follows:
\begin{equation}\label{eq-treedecompose}
    \mathbf{T}^k = \text{sign}(\widehat{\mathbf{A}}^{k}) - \text{sign}(\widehat{\mathbf{A}}^{k - 1}) + \mathbf{I},
\end{equation}
\begin{equation}\label{eq-sign}
    \text{sign}(\widehat{\mathbf{A}}^{k})_{ij} = \begin{cases}
    1, & \text{if $\widehat{\mathbf{A}}^{k}_{ij} > 0$}\\
    0, & \text{if $\widehat{\mathbf{A}}^{k}_{ij} = 0$},
    \end{cases}
\end{equation}
where $\widehat{\mathbf{A}}^{0} = \mathbf{I}$ is the identity matrix and $\widehat{\mathbf{A}}=\widetilde{\mathbf{D}}^{-\frac{1}{2}}\widetilde{\mathbf{A}}\widetilde{\mathbf{D}}^{-\frac{1}{2}}$ is the renormalized adjacency matrix as previously defined. The equivalence between $\mathbf{T}^{k}$ and the $k^{\text{th}}$-layer subgraph including the self-loop can be easily proven, so we omit the details for brevity.

\subsection{Multi-hop dependency}\label{sec-longrangedependency}
Although the tree decomposition could avoid the issue of feature smoothing between different layers, we also lose the multi-hop dependency captured by the original iterative propagation which might cause over-smoothing~\cite{GRAND}. Two nodes have multi-hop dependency if they are connected by a path in the network and specifically $k$-hop dependency is defined as two nodes that are connected by at least one simple path with length $k$. 
For example in Figure \ref{fig-tree-decomposition}, features of node $v_2$ cannot only be propagated along the edge $v_{2}\rightarrow v_{1}$ to $v_1$ but can also along the longer path $v_2\rightarrow v_5\rightarrow v_6 \rightarrow v_3 \rightarrow v_1$ to $v_1$. However after tree decomposition, the edge $v_2 \rightarrow v_1$ is the only way for propagating features of $v_2$ to $v_1$. Instead of inserting multiple edges between the higher-layer neighborhood nodes and their corresponding center nodes, we model this multi-hop dependency via graph diffusion~\cite{Diffusion}. Specifically, $\widehat{\mathbf{A}}^i$ represents the $i^{\text{th}}$-hop dependency and its entry $\widehat{\mathbf{A}}^i_{pq}$ measures the strength of paths of length $i$ in propagating features from node $v_p$ to $v_q$. Assuming the maximum hop of the dependency we consider is $K$, since $k^\text{th}$-layer ($k\le K$) neighborhood nodes can only propagate their features along paths of length from $k$ to $K$, thus the total multi-hop dependencies from node $v_p$ to $v_q$ along these paths is calculated as $\sum_{i = k}^{K}{\widehat{\mathbf{A}}^{i}}_{pq}$. Such multi-hop dependency across the spectrum from $k$ to $K$ between a single pair of nodes can be further generalized to all pairs of nodes in the graph via diffusion and defined as:
\begin{equation}\label{eq-diffusion}
    \mathbf{E}^{k, K} = \sum_{i = k}^{K}{\widehat{\mathbf{A}}^{i}},
\end{equation}
where $\mathbf{E}^{k, K}$ considers dependencies from paths of length $k$ to $K$, which could be used as the edge weights for propagating node features in the $k^\text{th}$-layer subgraph obtained from tree decomposition. 

\subsection{Tree Decomposed Graph Neural Network}
Now, having motivated and introduced the two major components of our proposed framework, namely the tree decomposition and multi-hop dependency formulations, we collect them together and present our Tree Decomposed Graph Neural Network (TDGNN). As previously noted, an illustration of our proposed TDGNN is shown in Figure \ref{fig-framework} and its corresponding mathematical formulation is defined as:
\begin{equation}\label{eq-MLP}
    \mathbf{H}^{0} = \text{MLP}(\mathbf{X}),
\end{equation}
\begin{equation}\label{eq-Prop}
    \mathbf{H}^{k} = (\mathbf{T}^{k}\odot \mathbf{E}^{k, K})\mathbf{H}^{0}, ~k = 1, 2, ..., L,
\end{equation}
\begin{equation}\label{eq-AGG}
    \mathbf{Z} = \begin{cases}
    \sum_{k = 0}^{L}{\mathbf{H}^{k}}, & \text{TDGNN-s}\\
    \sum_{k = 0}^{L}{\theta_{k}\mathbf{H}^{k}}, & \text{TDGNN-w.}
    \end{cases}
\end{equation}
\vspace{0.5ex}

We first apply a Multilayer Perceptron (MLP) network to the original feature matrix $\mathbf{X}$ to get the initial representations of nodes $\mathbf{H}^{0}$~\cite{APPNP, DAGNN}. Then, we decompose the whole network by calculating the adjacency matrix $\mathbf{T}^k$ of $k^\text{th}$-layer tree based on Eq.~\ref{eq-treedecompose} and Eq.~\ref{eq-sign}. This $k^\text{th}$-layer subgraph contains only edges between center nodes and their corresponding $k^\text{th}$-layer neighborhood nodes including a self-loop. Since here we consider the neighborhood nodes up to $L^\text{th}$-layer, the $k$ is from $1$ to $L$. Next we utilize graph diffusion to calculate the multi-hop dependency $\mathbf{E}^{k, K}$ based on Eq. \ref{eq-diffusion} and $K$ is the predefined maximum hop of dependency we consider. Afterwards, we propagate the initial node representations $\mathbf{H}^{0}$ along edges in each subgraph following each corresponding adjacency matrix $\mathbf{T}^{k}$ with the corresponding edge weight from multi-hop dependency $\mathbf{E}^{k, K}$ to get representations $\mathbf{H}^{k}$ for each layer $k$ based on Eq. \ref{eq-Prop}. We collect representations from each layer and aggregate them together using two aggregation mechanisms to get the final representations $\mathbf{Z}$ based on Eq. \eqref{eq-AGG}. The first aggregation mechanism is to directly sum up the representations of all layers together. The second aggregation mechanism is to assign learnable weights and adaptively combine the representations of each layer. The corresponding two versions of our model are termed as TDGNN-s and TDGNN-w, respectively. Ultimately, $\mathbf{Z}$ is employed to compute the cross-entropy loss for all labeled nodes as:
\begin{equation}
    \mathcal{L} = -\sum_{v_i\in\mathcal{V}_l}{\sum_{j = 1}^{C}{\mathbf{Y}_{ij}\log\widehat{\mathbf{Z}}_{ij}}},
\end{equation}
where $\widehat{\mathbf{Z}}$ is the probability distribution of each node belonging to each class and is obtained by applying softmax on the final representation $\mathbf{Z}$. Note that $\mathcal{V}_l \subset \mathcal{V}$ is the set of training nodes with known label information as previously defined and $C$ is the total number of classes to be predicted.

In summary, our model decouples representation transformation from propagation~\cite{DAGNN}, which enlarges the receptive fields without introducing more trainable parameters. Obtaining low dimensional representations before propagation follows is the idea of predicting than propagating~\cite{APPNP}, which makes the training process of TDGNN computationally efficient. Additionally, the tree decomposition preprocessing allows more flexible choice of utilizing and combining different layers to propagate features. The multi-hop dependency enables feature propagation along paths of various lengths, which conforms to other recent work~\cite{wang2020direct}. Furthermore, applying learnable weight coefficients equips TDGNN with the ability to flexibly select an effective receptive field based on a specific network and generate adaptive representations.

\subsection{Complexity Analysis}
In comparison to vanilla GCN, the additional computational load mostly comes from the tree decomposition and the graph diffusion. Since tracking down the corresponding $L$-layer subgraphs is equivalent to calculating the difference between corresponding powers of the normalized adjacency matrices with added self-loops by Eq. \ref{eq-treedecompose}, which is exactly given by graph diffusion, the time for the tree decomposition could be saved. The time complexity for performing graph diffusion process is $O(Kn^3) = O(n^3)$ due to $K$ times matrix multiplication and can be reduced approximately to $O(n^{2.81})$ if using the Strassen algorithm~\cite{strassen1969gaussian} or even further to $O(n^{2.38})$~\cite{coppersmith1987matrix}.
Moreover, in practice real-world graphs are extremely sparse and thus sparse matrix multiplication methods~\cite{yuster2005fast} could also be used to further improve computational efficiency. Notably, both of the tree decomposition and the graph diffusion are preprocessing outside of the training, which significantly reduces the computational load of the whole framework. 

For the space complexity, the bottleneck would be saving diffusion matrices $\widehat{\mathbf{A}}^k, k = 1, 2, ..., K$, and the adjacency matrices of each subgraph $\mathbf{T}^k, k = 1, 2, ..., K$, which leads to $O(Kn^2)$ and constitutes a severe threat for networks of large scale. However, as we highlighted before and demonstrate in Figure \ref{fig-layer}, since higher-layer neighborhoods may have completely different features from their corresponding center nodes and incorporating their information gain little benefits in learning better representation, we could only consider lower-layer neighborhoods~\cite{zeng2020deep} and thus keep only the first few adjacency matrices. On the other hand, most of the real-world networks have the small-world property that most nodes can be reached from every other node by a small number of hops~\cite{milgram1967small}, which confirms and helps justify why we can remove the higher-layer adjacency matrices. Moreover, we could apply the same strategy as GraphSAGE~\cite{Graphsage} where we sample nodes from center node's local neighborhood via random walk and propagate features among these sampled nodes~\cite{zeng2020deep}. Since this work mainly focuses on disentangling neighborhoods to avoid feature smoothing between different layers and characterizing the multi-hop dependency, the aforementioned is left as one future direction.
\begin{table*}[htbp!]
\footnotesize
\renewcommand{\arraystretch}{1}
\setlength{\extrarowheight}{1.25pt}
\setlength\tabcolsep{3pt}
\caption{Statistics of datasets.}
\vskip -2ex
\label{table-statist-dataset}
\centering
\begin{tabular}{p{15mm}ccccccc}
\hline
\multicolumn{2}{l}{\textbf{Networks}}          & \textbf{Nodes}     & \textbf{Edges}     & \textbf{Features}     & \textbf{Classes}     & \textbf{Train/Val/Test}  & \textbf{Type}     \\
\hline
\multirow{3}{*}{\textbf{Homophily}} &\multicolumn{1}{l}{Cora}& 2708 & 5429 & 1433 & 7      & \makecell{140/500/1000} &Citation network\\
& \multicolumn{1}{l}{Citeseer}& 3327 & 4732 & 3703 & 6 & 120/500/1000 &Citation network\\
& \multicolumn{1}{l}{Pubmed}& 19717 & 44338 & 500 & 3 & 60/500/1000 & Citation network\\
\hline
\multirow{4}{*}{\makecell{\textbf{Non-}\\\textbf{homophily}}} & \multicolumn{1}{l}{Cornell} & 183 & 295 & 1703 & 5 & 48\%/32\%/20\% & Webpage network\\
& \multicolumn{1}{l}{Texas} & 183 & 309 & 1703 & 5 & 48\%/32\%/20\% & Webpage network \\
& \multicolumn{1}{l}{Wisconsin} & 251 & 499 & 1703 & 5 & 48\%/32\%/20\% & Webpage network \\
& \multicolumn{1}{l}{Actor} & 7600 & 33544 & 931 & 5 & 48\%/32\%/20\% & Actor co-occurrence network \\
\hline
\end{tabular}
\end{table*}

\begin{table*}[]
\footnotesize
\renewcommand{\arraystretch}{1.1}
\setlength\tabcolsep{5pt}
\caption{Summary of semi-supervised classification accuracy (\%) $\pm$ stdev over Cora, Citeseer, and Pubmed datasets.}
\vskip -2ex
\label{table-semi}
\centering
\begin{tabular}{|l|cc|cc|cc|c|}
\hline
\multirow{2}{*}{\textbf{Method}} & \multicolumn{2}{c|}{\textbf{Cora}} & \multicolumn{2}{c|}{\textbf{Citeseer}} & \multicolumn{2}{c|}{\textbf{Pubmed}} & \multirow{2}{*}{\textbf{Avg. Rank}}  \\
\cline{2-7}
& Fixed & Random & Fixed & Random & Fixed & Random & \\
\hline
GCN & 81.50$\pm$0.79 (0-2) & 79.91$\pm$1.64 (0-2) & 71.42$\pm$0.48 (0-2) & 68.78$\pm$2.01 (0-2) & 79.12$\pm$0.46 (0-2) & 77.84$\pm$2.36 (0-2) & 7.17 \\
GAT & 83.10$\pm$0.40 (0-2) &  80.80$\pm$1.60 (0-2) &  70.80$\pm$0.50 (0-2) & 68.90$\pm$1.70 (0-2)  &  79.10$\pm$0.40 (0-2) & 77.80$\pm$2.10 (0-2)  &  7.00 \\
SGC & 82.63$\pm$0.01 (0-2) & 80.18$\pm$1.57 (0-2) & 72.10$\pm$0.14 (0-2) & 69.33$\pm$1.90 (0-2) & 79.12$\pm$0.10 (0-2) & 76.74$\pm$2.84 (0-2)& 6.83\\
APPNP & 83.34$\pm$0.56 (0-10) & 82.26$\pm$1.39 (0-10) & 72.22$\pm$0.50 (0-10) & 70.53$\pm$1.57 (0-10) & 80.14$\pm$0.24 (0-10) & 79.54$\pm$2.23 (0-10) & 3.83\\
DAGNN & 84.88$\pm$0.49 (0-10) & 83.47$\pm$1.18 (0-10) & 73.39$\pm$0.57 (0-9) & 70.87$\pm$1.44 (0-10) & \textbf{80.51$\pm$0.42 (0-20)} & 79.52$\pm$2.19 (0-20) & 2.33\\
GCNII* &  \textbf{85.57}$\pm$\textbf{0.45 (0-64)}& 82.58$\pm$1.68 (0-64) & 73.24$\pm$0.61 (0-32) &70.04$\pm$1.72 (0-10) & 80.00$\pm$0.48 (0-16)& 79.03$\pm$1.68 (0-16) & 3.83\\
TDGNN-s & 85.35$\pm$0.49 (0-4)  & \textbf{83.84}$\pm$\textbf{1.45 (0-6)} &    \textbf{73.78}$\pm$\textbf{0.60 (0-8)}  &  \textbf{71.27}$\pm$\textbf{1.71 (0-8)} & 80.20$\pm$0.33 (0-5)& \textbf{80.01}$\pm$\textbf{1.96 (0-5)} & \textbf{1.33} \\
TDGNN-w & 84.42$\pm$0.59 (0-4)  & 83.43$\pm$1.35 (0-6) & 72.14$\pm$0.49 (0-6) & 70.32$\pm$1.57 (0-6) & 80.12$\pm$0.44 (0-5) &79.77$\pm$2.04 (0-5) & 3.67\\
\hline
\end{tabular}
\vskip -1ex
\end{table*}

\section{Experiments}\label{sec-experiments}
In this section, we conduct extensive node classification experiments to evaluate the superiority of our proposed TDGNN model. We begin by introducing the datasets and experimental setup we employed. Then, we compare TDGNN with prior baselines and some state-of-the-art (SOTA) deep GNNs. 

\subsection{Experimental Settings}

\subsubsection{Datasets}
We evaluate the performance of our TDGNN model and baseline models with node classification on multiple real-world datasets. More specifically, we use the three standard citation network datasets Cora, Citeseer, and Pubmed~\cite{sen2008collective} for semi-supervised node classification~\cite{yang2016revisiting}, where nodes correspond to documents associated with the bag-of-words as the features and edges correspond to citations. For full-supervised node classification, in addition to the three citation networks we include three extra web network datasets, Cornell, Texas, and Wisconsion~\cite{geomgcn}, where nodes and edges represent web pages and hyperlinks, and one actor co-occurrence network dataset, Actor~\cite{geomgcn}, where nodes and edges represent actors and their co-occurrence in the same movie.
Table~\ref{table-statist-dataset} contains the basic network statistics for each of these datasets.

\subsubsection{Baselines}
To evaluate the effectiveness of TDGNN, we choose the following representative supervised node classification baselines including SOTA GNN models.
\begin{itemize}[leftmargin=0.6cm]

    \item \textbf{MLP}~\cite{murtagh1991multilayer}: 2-layer multilayer perceptron with dropout and ReLU non-linearity, which is empirically shown in other works to perform well on non-homophily network datasets~\cite{zhu2020beyond}.
    
    \item \textbf{GCN}~\cite{GCN}: GCN is one of the most popular graph convolutional models and our proposed model is modified based on it.
    
    \item \textbf{GAT}~\cite{GAT}: Graph attention network employs attention mechanism to pay different levels of attention to nodes within the neighorhood set, and is widely used as a GNN baseline. 
    
    \item \textbf{SGC}~\cite{SGC}: Simple graph convolution network removes nonlinearities and collapsing weight matrices between consecutive layers, which obtains the comparable accuracy and yields orders of magnitude speedup over GCN. We note that SGC collapses the traditional GNN aggregation tree such that the center node receives the features directly from the flattened neighborhood while being weighted according to the higher-order neighborhood information.
    
    \item \textbf{APPNP}~\cite{APPNP}: APPNP links GCN and PageRank to derive an improved propagation scheme based on personalized PageRank, which incorporates higher-order neighborhood information and meanwhile keeps the local information.
    
    \item \textbf{Geom-GCN}~\cite{geomgcn}: Geom-GCN explores to capture long-range dependencies in non-homophily networks. It uses the geometric relationships defined in the latent space to build structural neighorhoods for aggregation. Since Geom-GCN is mainly designed for non-homophily networks, we only report its performance in full-supervised node classification where three non-homophily networks are included.
    
    \item \textbf{DAGNN}~\cite{DAGNN}: Deep adaptive graph neural network first decouples the representation transformation from propagation so that large receptive fields can be applied without suffering from performance degradation. Then, it utilizes an adaptive adjustment mechanism, which adaptively balances the information from local and global neighborhoods for each node.
    
    \item \textbf{GCNII}~\cite{GCNII}: GCNII employs residual connection to retain part of the information from the previous layer and adds an identity mapping to ensure the non-decreasing performance as the GNN model goes deeper (i.e., successfully adds more layers).
\end{itemize}

For baselines that have multiple variants (Geom-GCN, GCNII), we only choose the best for each dataset and denote it as model*.

\begin{table*}[t]
\footnotesize
\renewcommand{\arraystretch}{1.1}
    \setlength\tabcolsep{3pt}
\caption{Summary of full-supervised classification accuracy (\%) $\pm$ stdev over 8 datasets.}
\vskip -2ex
\label{table-full}
\centering
\begin{tabular}{|l|ccccccc|c|}
\hline
\textbf{Method} & \textbf{Cora} & \textbf{Cite.} & \textbf{Pub.} & \textbf{Corn.}& \textbf{Tex.}& \textbf{Wisc.}& \textbf{Act.}& \textbf{Avg. Rank}\\
\hline
MLP &75.78$\pm$1.84 (0) & 73.81$\pm$ 1.74 (0)& 86.90$\pm$0.37 (0)& 80.97$\pm$6.33 (0) & 81.32$\pm$ 4.19 (0) & 85.38$\pm$3.95 (0) & 36.60$\pm$1.25 (0) & 5.57\\
GCN & 86.97$\pm$1.32 (0-2) & 76.37$\pm$1.47 (0-2) & 88.19$\pm$0.48 (0-2) & 58.57$\pm$3.57 (0-2) & 58.68$\pm$4.64 (0-2) &53.14$\pm$6.25 (0-2) & 28.65$\pm$1.38 (0-2)& 8.14\\
GAT & 87.30$\pm$1.01 (0-2) & 75.55$\pm$1.32 (0-2)& 85.33$\pm$0.48 (0-2)& 61.89$\pm$5.05 (0-2)& 58.38$\pm$6.63 (0-2)& 55.29$\pm$4.09 (0-2)& 28.45$\pm$0.89 (0-2)& 8.00\\
SGC & 87.07$\pm$1.20 (0-2) & 76.01$\pm$1.78 (0-2) & 85.11$\pm$0.52 (0-2) & 58.68$\pm$3.75 (0-2)&  60.43$\pm$5.11 (0-2)& 53.49$\pm$5.13 (0-2)& 27.46$\pm$1.46 (0-2)& 8.57\\
Geom-GCN* & 85.35$\pm$1.57 (0-2) & \textbf{78.02$\pm$1.15 (0-2)} & 89.95$\pm$0.47 (N/A) &  60.54$\pm$3.67 (0-2) &   66.76$\pm$2.72 (N/A)  &64.51$\pm$3.66 (N/A)& 31.63$\pm$1.15 (N/A)& 5.86\\
APPNP & 86.76$\pm$1.74 (0-10) & 77.08$\pm$1.56 (0-10)& 88.45$\pm$0.42 (0-10)& 74.59$\pm$5.11 (0-10)& 74.30$\pm$4.74 (0-10)&81.10$\pm$2.93 (0-10)& 34.36$\pm$1.09 (0-10)& 5.43\\
DAGNN &  87.26$\pm$1.42 (0-10) & 76.47$\pm$1.54 (0-10)& 87.49$\pm$0.63 (0-20) & 80.97$\pm$6.33 (0)& 81.32$\pm$4.19 (0) & 85.38$\pm$3.95 (0) & 36.60$\pm$1.25 (0) & 4.71\\
GCNII* &  \textbf{88.27}$\pm$\textbf{1.31 (0-64)} & 77.06$\pm$1.67 (0-64) & \textbf{90.26$\pm$0.41 (0-64)} & 76.70$\pm$5.40 (0-16) &77.08$\pm$5.84 (0-32) & 80.94$\pm$4.94 (0-16) & 35.18$\pm$1.30 (0-64) & 3.71\\
TDGNN-s & 88.26$\pm$1.32 (0-4) & 76.64$\pm$1.54 (0-8) & 89.13$\pm$0.39 (0-1) & 80.97$\pm$6.33 (0)& 82.95$\pm$4.59 (0, 4-5)& 85.47$\pm$3.88 (0, 4-5) & 36.70$\pm$1.28 (0, 3-4)& 2.86\\
TDGNN-w & 88.01$\pm$1.32 (0-5)& 76.58$\pm$1.40 (0-2)& 89.22$\pm$0.41 (0-1)& \textbf{82.92$\pm$6.61 (0, 2-6)} & \textbf{83.00$\pm$4.50 (0, 2)} & \textbf{85.57$\pm$3.78 (0, 3-5)}& \textbf{37.11$\pm$0.96 (0, 3-4)} & \textbf{2.14}\\
\hline
\end{tabular}
\begin{tablenotes}
      \small
      \centering
      \item \textbf{*} We reuse the results reported in~\cite{coin} for Geom-GCN. 'N/A' indicate the corresponding layers are not reported in the paper.
\end{tablenotes}
\vspace{-1.5em}
\end{table*}

\subsubsection{Parameter Settings}
We implement our proposed TDGNN and some necessary baselines using Pytorch~\cite{pytorch} and Pytorch Geometric~\cite{paszke2019pytorch}, a library for deep learning on graph-structured data built upon Pytorch. For DAGNN\footnote{\url{https://github.com/vthost/DAGNN}}, and GCNII\footnote{\url{https://github.com/chennnM/GCNII}}, we use the original code from the authors' github repository. We aim to provide a rigorous and fair comparison between different models on each dataset by tuning hyperparameters for all models individually. The number of hidden unit is searched from $\{16, 32, 64, 128\}$, the dropout rate is searched from $\{0, 0.5, 0.8\}$, the weight decay is searched from $[1e^{-4}, 2e^{-2}]$, the training epochs is searched from $\{300, 500, 1000, 1500, 3000, 4000\}$ and the learning rate is set to be 0.01. We find that some baselines even achieve better results than their original reports. Note that in this work, we do not treat the random seed as a hyperparamter and therefore, the random seed fixed in previous models for reproducing results, if any, is reset to be totally random to remove any potential bias and thus allow for more generalized comparison. For reproducibility, codes of all of our models and corresponding hyperparameter configurations for results in Table \ref{table-semi}-\ref{table-full} are publicly available \footnote{\url{https://github.com/YuWVandy/TDGNN}\label{github}}. 

\subsection{Semi-supervised Node Classification}
For the semi-supervised node classification task, we apply the fixed split following~\cite{yang2016revisiting} and random training/validation/testing split on Cora, Citeseer, and Pubmed, with 20 nodes per class for training, 500 nodes for validation and 1000 nodes for testing. For each model, we conduct 100 runs and report the mean classification accuracy with the standard deviation in both the fixed and random splitting cases. Table \ref{table-semi} reports the best mean accuracy with the standard deviation over different data splits where the best model per benchmark is highlighted in bold and the number in parentheses corresponds to layers of neighborhoods utilized at which the best performance is achieved. For example, (0-4) means the corresponding performance is achieved when we use neighborhood of layers up to 4 and $0$-layer neighborhoods correspond to using center nodes themselves.

We observe that TDGNN-s performs the best in terms of the average rank through all datasets and across both random and fixed splits, which suggests the comprehensive superiority of TDGNN-s to other baselines. Specifically, our TDGNN-s model outperforms the representative baselines including GCN, GAT, SGC, and APPNP across all datasets by significant margins. Compared with two recent deep GNN models, DAGNN and GCNII*, TDGNN-s can still achieve the comparable or even better performance. Especially when the data split is random, TDGNN-s outperforms all other models, which demonstrates the strong robustness of TDGNN-s (in terms of dataset splits). It is also worthwhile to note that our TDGNN model achieves the SOTA performance with relatively shallow layers compared with DAGNN and GCNII*. On Cora dataset, the best performance is achieved when layers are used up to 4 and 6 for our TDGNN-s model, respectively, in fixed and random data splitting, while DAGNN and GCNII require up to 10 and 64 layers to achieve the best, which demands heavy computation and thus are time inefficient. On Citeseer dataset, our model also utilizes up to the most shallow layers compared with DAGNN and GCNII* to achieve the SOTA performance. Surprisingly, the weighted version of our model, TDGNN-w, has poorer performance than TDGNN-s while still outperforms most of the baselines. This is because the weight coefficients $\{\theta_0, \theta_1, ..., \theta_L\}$ are only decided by training nodes and the suitable weights for combining aggregated features $\{\mathbf{H}_{0}, \mathbf{H}_{1}, ..., \mathbf{H}_{L}\}$ and getting good predictions on training and validation nodes might not be suitable for testing nodes, which inspires future work for a layer aggregation mechanism that enables node-adaptive layer combination.

\vspace{-0.75ex}
\subsection{Full-supervised Node Classification}
For the full-supervised node classification task, we evaluate our TDGNN model and existing GNNs using 7 datasets: Cora, Citeseer, Pubmed, Cornell, Texas, Wisconsin, and Actor. For each dataset, we use 10 random splits (48\%/32\%/20\% of nodes per class for training/validation/testing) from~\cite{geomgcn}\footnote{Note that although~\cite{geomgcn} reports that the ratios are 60\%/20\%/20\%, but this is different from the actual data splits shared on their GitHub~\cite{coin}.}. We conduct 100 runs with each split evaluated 10 times and report the mean accuracy with the standard deviation in Table \ref{table-full}. We note that here the numbers in the parentheses again correspond to layers of neighborhoods utilized (e.g., (0,3-5) means the corresponding performance is achieved when we use neighborhoods of layers 3 to 5 and $0$-layer neighborhoods corresponding to the center nodes themselves.) 

First, we observe from Table~\ref{table-full} that TDGNN-w has the best average rank across the two types of networks (i.e., homophily and heterophily) with TDGNN-s ranks second. Next, we observe that TDGNN-w significantly outperforms the baselines across the heterophily networks. However, both variants of TDGNN are slightly outperformed on the homophily networks in this full-supervised setting (whereas in most homophily networks under the semi-supervised setting TDGNN-s performs the best). Thus, to better understand the inner workings of TDGNN, we next perform a detailed parameter analysis.

\vspace{-0.75ex}
\subsection{Parameter Analysis}
Here we compare the performance of TDGNN with other baselines when utilizing neighborhoods in different layers. Furthermore, we perform a parameter analysis of TDGNN by varying the neighborhood layers ($L$) and the multi-hop dependencies ($K$).

\begin{figure}[t]
     \centering
     \hspace{-1.5ex}
     \begin{subfigure}[b]{0.235\textwidth}
         \centering
         \includegraphics[width=1.02\textwidth]{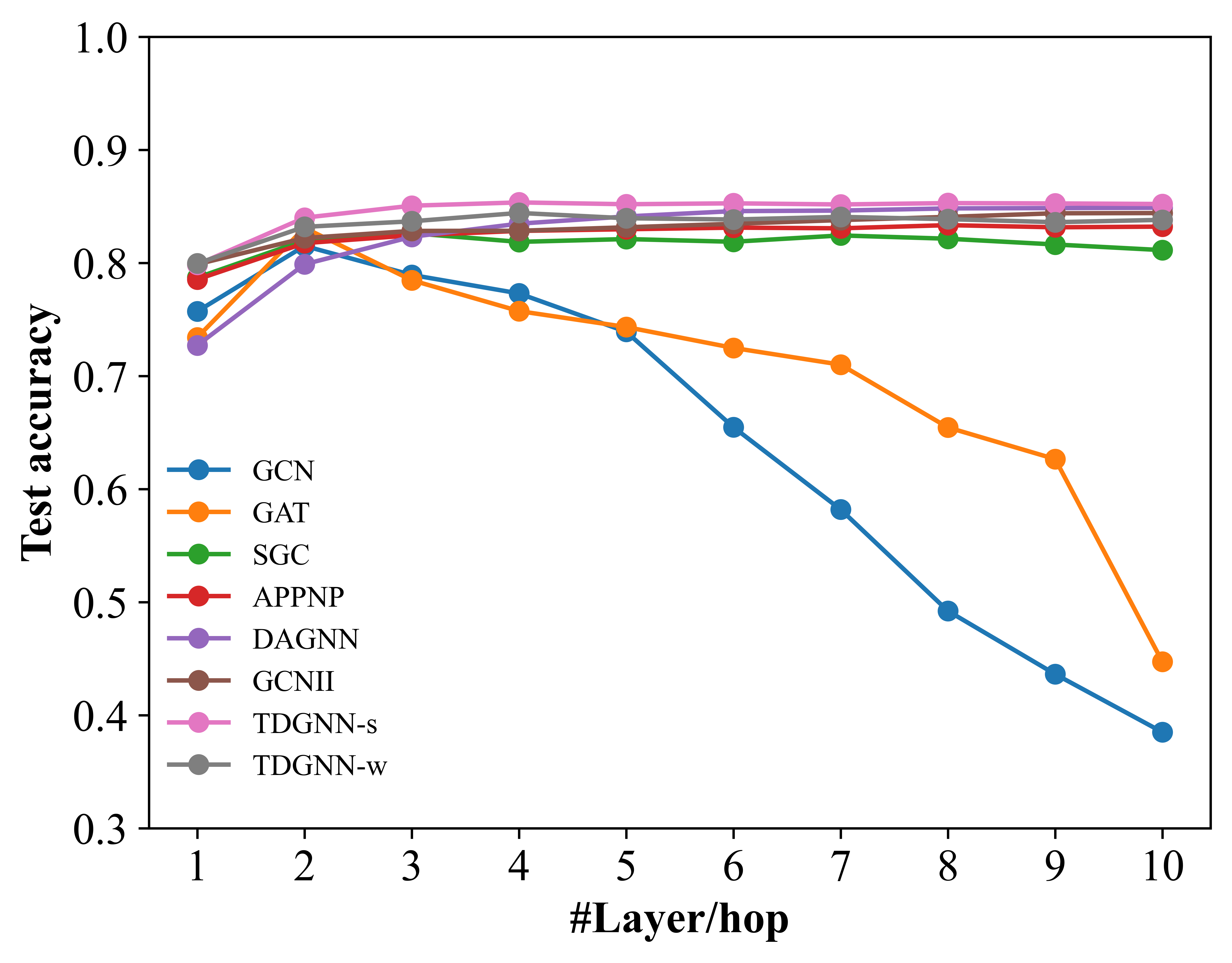}
         \vskip -0.25ex
         \caption{Cora}
         \label{fig-layercora}
     \end{subfigure}
     \hspace{-1ex}
     \begin{subfigure}[b]{0.235\textwidth}
         \centering
         \includegraphics[width=1.02\textwidth]{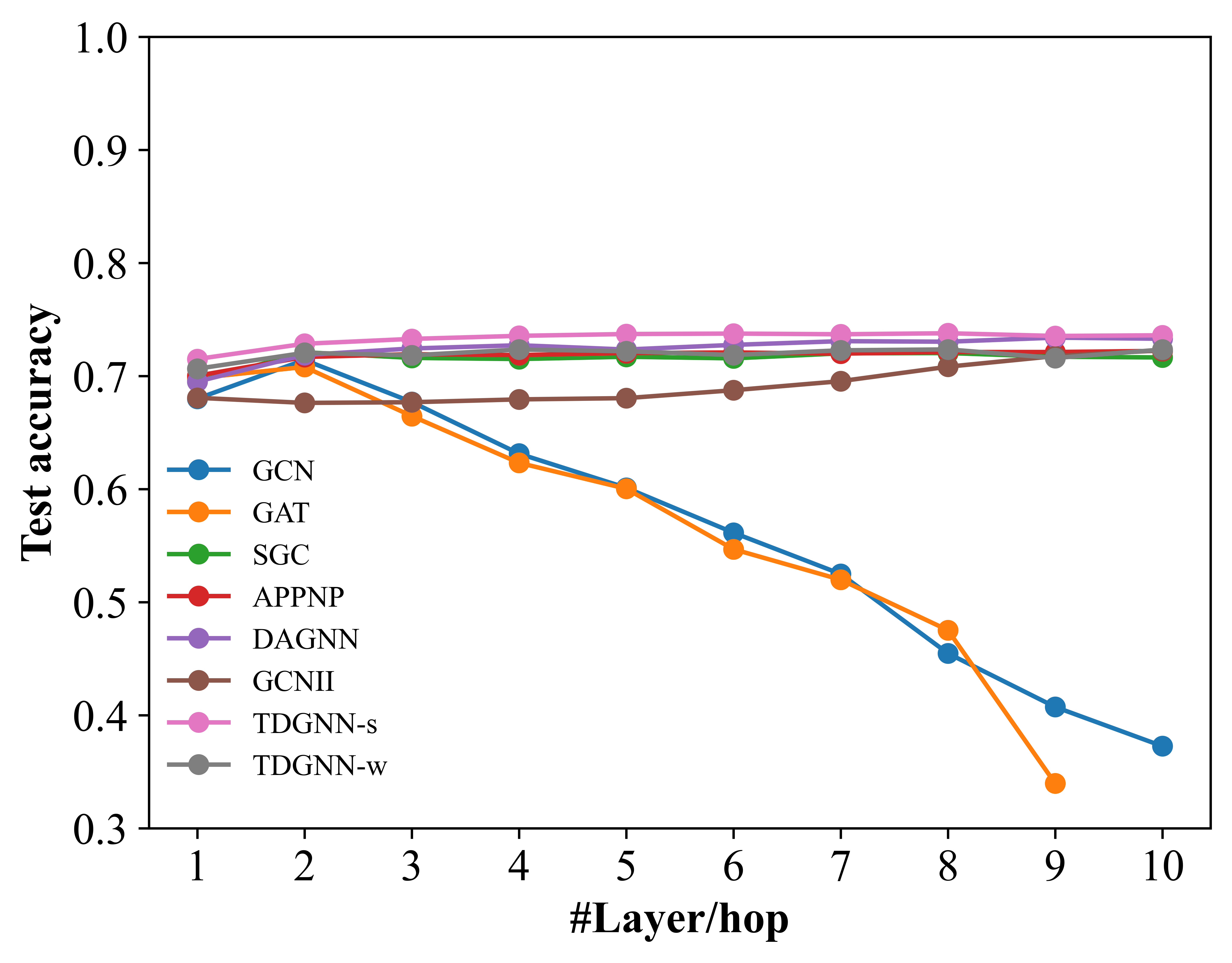}
         \vskip -0.25ex
         \caption{Citeseer}
         \label{fig-layerciteseer}
     \end{subfigure}
     \vskip -2ex
     \caption{Results of models with different layers.}
     \label{fig-layer}
     \vskip -3.5ex
\end{figure}

First, to demonstrate the strength of the TDGNN-s model in shallow layers, we visualize the performance of each model using layers from up to 1 to up to 10 in Figure~\ref{fig-layer}. For the Cora and Citeseer datasets, our model achieves around 84\% and 73\% using only the first 2-layer neighborhoods and the first three-layer neighborhoods, respectively. Compared to two SOTA deep GNNs where DAGNN achieves the same level performance using 5 layers and 7 layers, and GCNII* achieves using 8 layers and at least 32 layers, our model can leverage less neighborhood information to achieve comparable performance, which clearly validates the importance of considering multi-hop dependency. This further to some extent raises the concern over whether we need deep GNNs to incorporate higher-layer neighborhood information in homophily networks, or if shallow feature information aggregated according to higher-order multi-hop dependencies provides sufficient information. Besides, the continued high-level performance as model depth increases demonstrates the higher resilience of TDGNN-s against over-smoothing.

\begin{figure}[t]
     \centering
     \begin{subfigure}[b]{0.245\textwidth}
         \centering
         \includegraphics[width=0.9\textwidth]{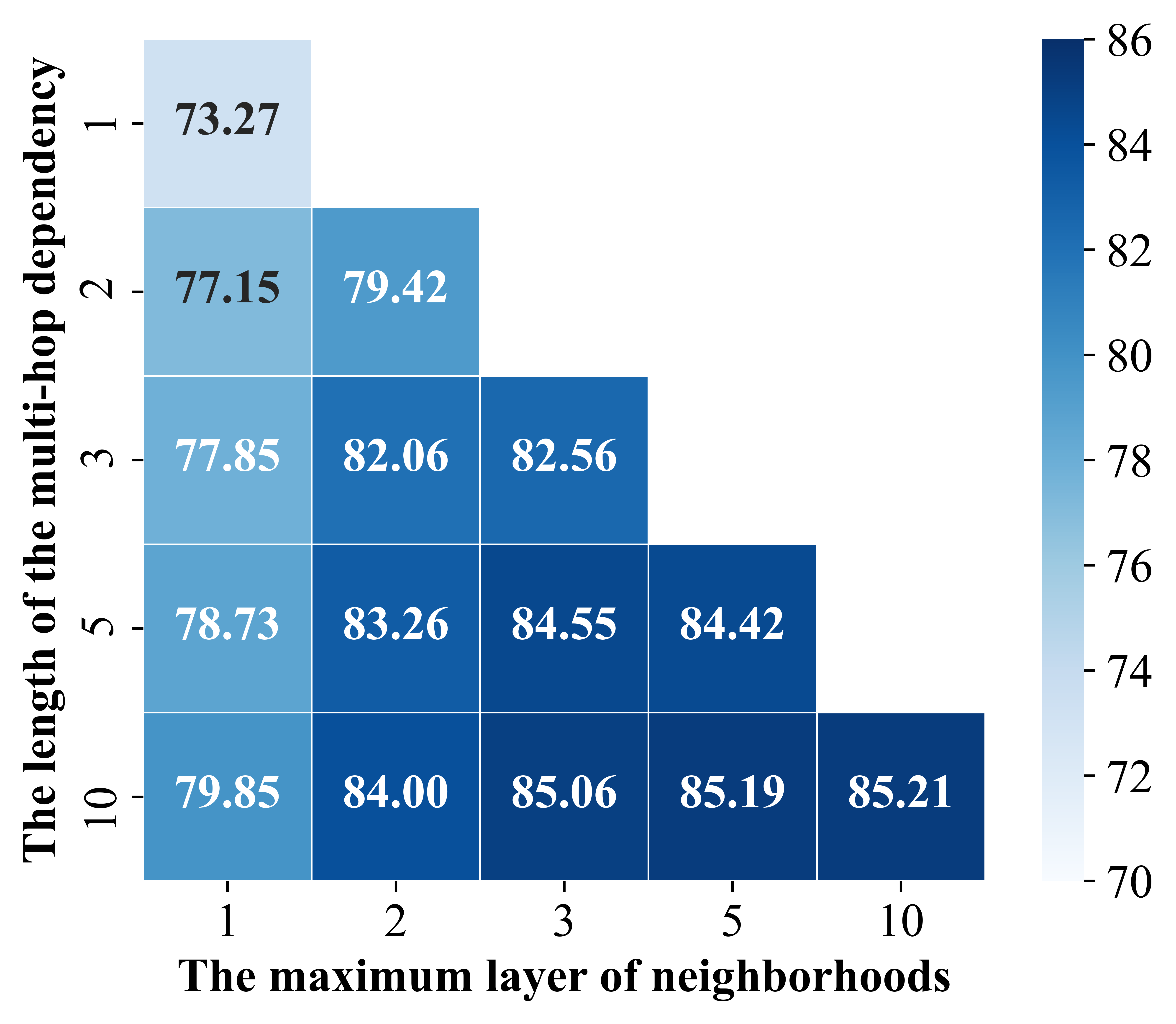}
         \vskip -1ex
         \caption{Cora (TDGNN-s)}
         \label{fig-coraheat}
     \end{subfigure}
     \hspace{-3ex}
     \begin{subfigure}[b]{0.245\textwidth}
         \centering
         \includegraphics[width=0.9\textwidth]{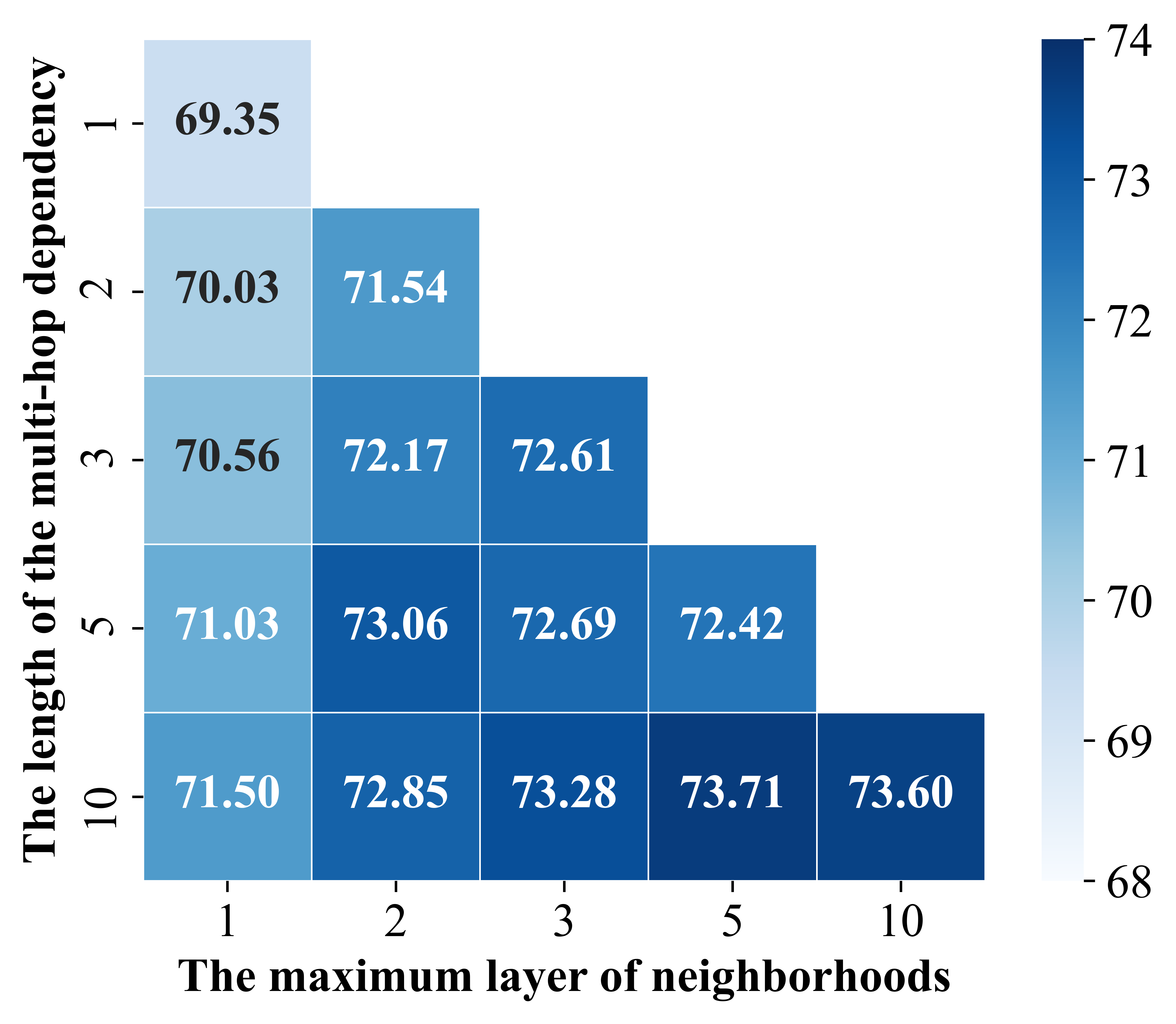}
         \vskip -1ex
         \caption{Citeseer (TDGNN-s)}
         \label{fig-citeheat}
     \end{subfigure}
     \begin{subfigure}[b]{0.245\textwidth}
         \centering
         \includegraphics[width=0.9\textwidth]{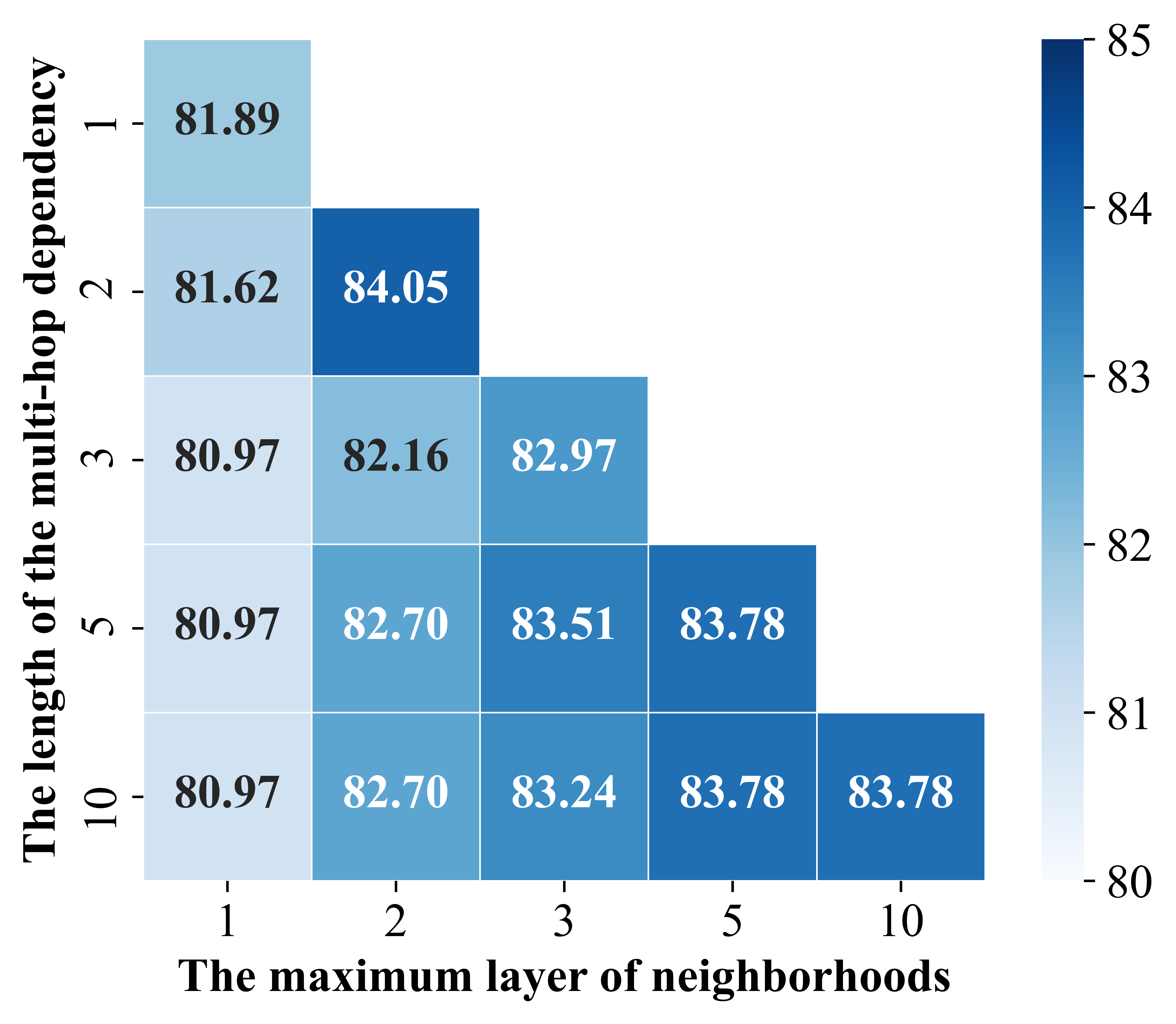}
         \vskip -1ex
         \caption{Cornell (TDGNN-w)}
         \label{fig-cornellheat}
     \end{subfigure}
     \hspace{-3ex}
     \begin{subfigure}[b]{0.245\textwidth}
         \centering
         \includegraphics[width=0.9\textwidth]{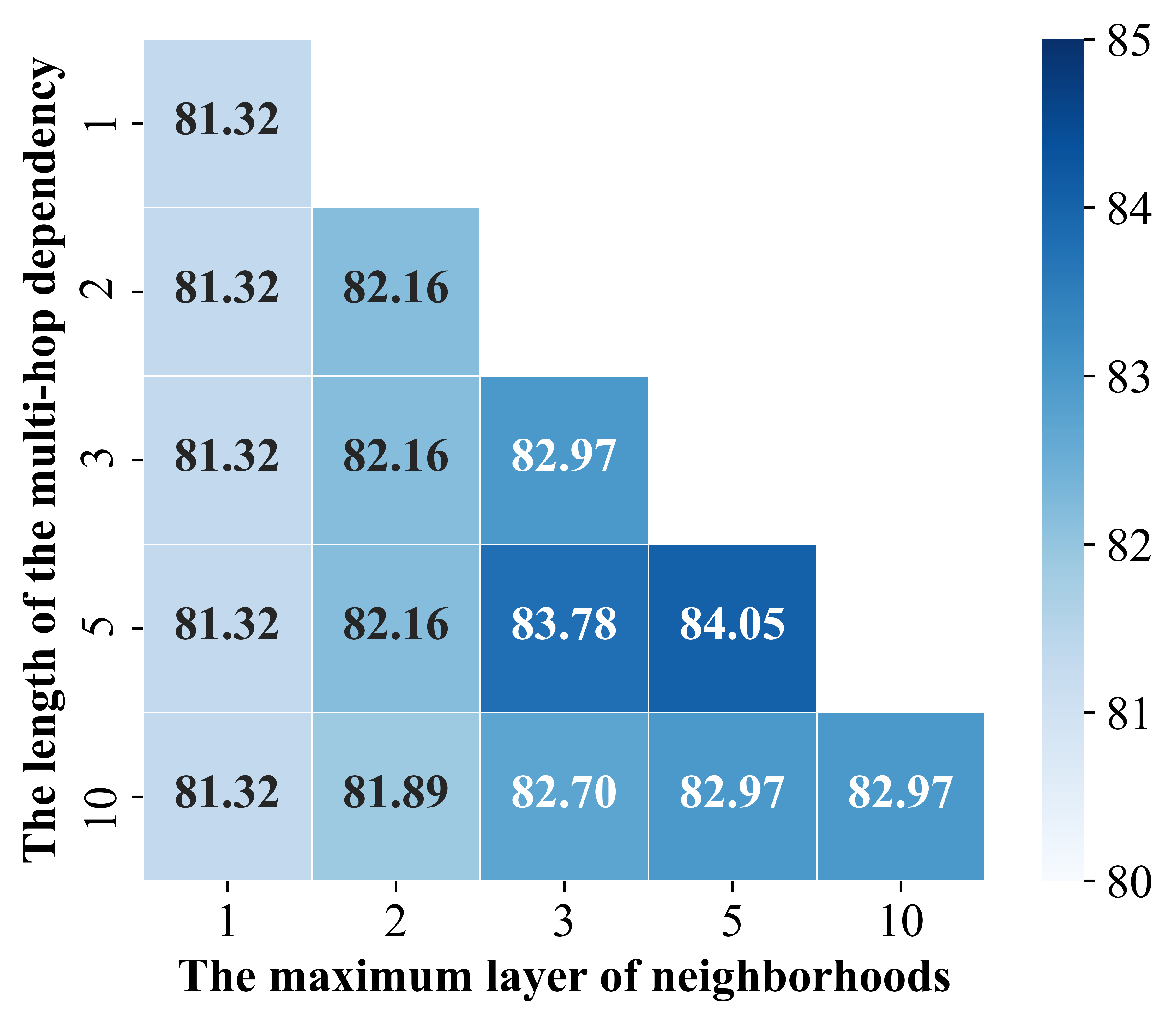}
         \vskip -1ex
         \caption{Texas (TDGNN-w)}
         \label{fig-texasheat}
     \end{subfigure}
     \vskip -1.25ex
     \caption{Visualizing the effect of varying the maximum layer neighborhoods and the length of mutli-hop dependency on the performance of TDGNN.}
     \label{fig-heath}
     \vskip -4ex
\end{figure}

Second, we vary the maximum layer of neighborhoods and the multi-hop dependency to study their effect on the performance of the proposed two models: TDGNN-s on two representative homophily networks and TDGNN-w on two representative heterophily networks. Both of the maximum layer of the neighborhoods and the length of the multi-hop dependency are selected from $\{1, 2, 3 ,5, 10\}$ due to the small-world theory that two nodes will be connected through few series of intermediaries~\cite{milgram1967small}. Figure~\ref{fig-heath} visualizes the averaged accuracy across 10 runs for various layers and dependency configurations. For two homophily networks, including extra neighborhood layers significantly increase the model performance for lower-layers and such boosting effect becomes progressively weaker as more and more higher-layer neighborhood layers are included, e.g., the performance increases from $79.85$ to $84.00$ and from $71.50$ to $72.85$ for Cora and Citeseer when including the $2^\text{nd}$-layer neighborhood while only from $84.00$ to $85.06$ and from $72.85$ to $73.28$ when including the $3^\text{rd}$-layer. This weaker boost as the layer number increases is also in line with the decreasing homophily level as observed in Figure~\ref{fig-distcora}. In comparison, for heterophily networks, in Figures~\ref{fig-cornellheat} and~\ref{fig-texasheat} we can observe a more significant need for the decoupling of neighborhood layers since increasing the receptive field (i.e., increasing the maximum layer of neighborhoods) is not always advantageous. Similarly including deeper multi-hop dependencies is not always a clear advantage as seen in the homophily networks because lower-layer neighborhoods that have different labels or representations from their corresponding center nodes may contribute more to their center nodes' prediction through longer dependency. We note that these findings also align with our empirical analysis in Figure~\ref{fig-distcora}.
Therefore, we believe that the increased performance obtained by TDGNN over prior work is partially credited to its ability to separate the concept of graph convolutions in deeper GNNs with higher-layer neighborhoods into both multi-hop dependencies and decoupled neighborhood layers, which can allow any deep GNN model to be more flexibly customized via hyperparameter tuning on a wider variety of complex networks.

\vspace{-1.5ex}
\section{Related Work}\label{sec-relatedwork}
Our work is related to previous work attempting to tackle the over-smoothing problem in GNNs (especially in deeper GNNs), and also related to prior work attempting to alleviate the challenges faced when applying GNNs to non-homophily graphs.

\vspace{-1.5ex}
\subsection{\hspace{-0.5ex}Over-smoothing Problem and Deep GNNs}
Over-smoothing derives from stacking multiple propagation layers resulting in feature information of nodes among different classes becoming indistinguishable. Previous work~\cite{zhou2020graph, li2018deeper} have proven that over-smoothing is a common phenomenon in many GNNs and smoothness among nodes from the same class is helpful for node classification. In~\cite{deng2019batch}, they discovered that noisy topology information results in feature over-smoothing and may lead to node misclassification. The smoothing in GNNs was later classified into two kinds by the information-to-noise ratio~\cite{Delioversmoothing}.
In this work, we study the feature smoothing between different layers, show 
that such type of smoothing inevitably happens when employing an iterative propagation framework and that it can be solved by tree decomposition.

Deep GNNs are related to over-smoothing and have the primary goal to incorporate higher-layer neighborhood information through iterative propagation.
For example, SGC~\cite{SGC} and S$^{2}$GC~\cite{SGCC} attempt to capture higher-layer neighborhood information by applying $K^\text{th}$ power of the graph convolution in a single neural network layer. APPNP~\cite{APPNP} replaces the power of the graph convolution with the Personalized PageRank~\cite{page1999pagerank} and GDC~\cite{Diffusion} further extends APPNP by generalizing Personalized PageRank to an arbitrary graph diffusion process. There are also more recent methods, such as GCNII~\cite{GCNII} and DAGNN~\cite{DAGNN}, which we have used as baselines for TDGNN since they have outperformed previously mentioned deep GNNs.

\vspace{-1.5ex}
\subsection{GNNs on Heterophily Networks}
Heterophily has recently been raised as an important issue since it breaks the traditional network homophily assumption that is widely adopted in many GNNs. More specifically, in a heterophily (i.e., non-homophily) network, the concept that linked nodes are likely from different classes or have dissimilar features is initially recognized within the context of GNNs in~\cite{geomgcn}. Zhu et al.~\cite{zhu2020beyond} proposes a set of effective designs that allow GNNs to generalize to challenging heterophily settings, and Chen et al.~\cite{GCNII} leverages initial residual connection and identity mapping to enable GCN to express a $K^\text{th}$ order polynomial filter with arbitrary coefficients, which achieves great progress in both homophily and heterophily networks. In comparison, our work demonstrates that the poor performance of GNNs on heterophily networks is caused by feature smoothing between neighborhoods in different layers. By decomposing the computational tree of center nodes and increasing the depth of the GNNs, we can selectively devise suitable layer configurations to boost the model performance on heterophily network.

\vspace{-1ex}
\section{Conclusion}\label{sec-conclusion}
In this paper, we theoretically analyze the feature smoothing of neighborhoods in different layers and propose a tree decomposition method that disentangles neighborhoods of different layers and thus allows more flexible layer configuration. Moreover, our work provides the first theoretical and empirical analysis that unveils the importance of multi-hop dependency in learning better node representations and discloses its connection with graph diffusion. Based on these insights, we design our Tree Decomposed Graph Neural Network (TDGNN) model
with two variants, TDGNN-s and TDGNN-w, which simultaneously address the problem of feature smoothing between different layers and incorporate the multi-hop dependency. Extensive experiments demonstrate that TDGNN outperforms representative baselines on a wide range of real-world datasets across network types (including homophily and heterophily) and various node classification task settings. 

For future work, we plan to devise a node-adaptive layer aggregation mechanism which can optimize the configurations of representations from different layers in a node specific way.
Such optimization could be realized by applying policy-based reinforcement learning. Furthermore, self-supervised learning (SSL) could be utilized to pre-train the MLP in our framework, which could enhance the capability of our model to embed more useful feature and topology information from diverse datasets, since SSL has recently been shown effective on GNNs~\cite{SelfTask,GNNBook-ch18-wang}.


\bibliographystyle{ACM-Reference-Format}
\bibliography{references}


\begin{thebibliography}{45}


\ifx \showCODEN    \undefined \def \showCODEN     #1{\unskip}     \fi
\ifx \showDOI      \undefined \def \showDOI       #1{#1}\fi
\ifx \showISBNx    \undefined \def \showISBNx     #1{\unskip}     \fi
\ifx \showISBNxiii \undefined \def \showISBNxiii  #1{\unskip}     \fi
\ifx \showISSN     \undefined \def \showISSN      #1{\unskip}     \fi
\ifx \showLCCN     \undefined \def \showLCCN      #1{\unskip}     \fi
\ifx \shownote     \undefined \def \shownote      #1{#1}          \fi
\ifx \showarticletitle \undefined \def \showarticletitle #1{#1}   \fi
\ifx \showURL      \undefined \def \showURL       {\relax}        \fi
\providecommand\bibfield[2]{#2}
\providecommand\bibinfo[2]{#2}
\providecommand\natexlab[1]{#1}
\providecommand\showeprint[2][]{arXiv:#2}

\bibitem[\protect\citeauthoryear{Cai and Ji}{Cai and Ji}{2020}]%
        {multilink}
\bibfield{author}{\bibinfo{person}{Lei Cai} {and} \bibinfo{person}{Shuiwang
  Ji}.} \bibinfo{year}{2020}\natexlab{}.
\newblock \showarticletitle{A Multi-Scale Approach for Graph Link Prediction}.
  In \bibinfo{booktitle}{\emph{The Thirty-Fourth {AAAI} Conference on
  Artificial Intelligence, {AAAI} 2020, The Thirty-Second Innovative
  Applications of Artificial Intelligence Conference, {IAAI} 2020, The Tenth
  {AAAI} Symposium on Educational Advances in Artificial Intelligence, {EAAI}
  2020, New York, NY, USA, February 7-12, 2020}}.
\newblock


\bibitem[\protect\citeauthoryear{Chen, Lin, Li, Li, Zhou, and Sun}{Chen
  et~al\mbox{.}}{2020a}]%
        {Delioversmoothing}
\bibfield{author}{\bibinfo{person}{Deli Chen}, \bibinfo{person}{Yankai Lin},
  \bibinfo{person}{Wei Li}, \bibinfo{person}{Peng Li}, \bibinfo{person}{Jie
  Zhou}, {and} \bibinfo{person}{Xu Sun}.} \bibinfo{year}{2020}\natexlab{a}.
\newblock \showarticletitle{Measuring and Relieving the Over-Smoothing Problem
  for Graph Neural Networks from the Topological View}. In
  \bibinfo{booktitle}{\emph{The Thirty-Fourth {AAAI} Conference on Artificial
  Intelligence, {AAAI} 2020, The Thirty-Second Innovative Applications of
  Artificial Intelligence Conference, {IAAI} 2020, The Tenth {AAAI} Symposium
  on Educational Advances in Artificial Intelligence, {EAAI} 2020, New York,
  NY, USA, February 7-12, 2020}}.
\newblock


\bibitem[\protect\citeauthoryear{Chen, Wei, Huang, Ding, and Li}{Chen
  et~al\mbox{.}}{2020b}]%
        {GCNII}
\bibfield{author}{\bibinfo{person}{Ming Chen}, \bibinfo{person}{Zhewei Wei},
  \bibinfo{person}{Zengfeng Huang}, \bibinfo{person}{Bolin Ding}, {and}
  \bibinfo{person}{Yaliang Li}.} \bibinfo{year}{2020}\natexlab{b}.
\newblock \showarticletitle{Simple and Deep Graph Convolutional Networks}. In
  \bibinfo{booktitle}{\emph{Proceedings of the 37th International Conference on
  Machine Learning, {ICML} 2020}}.
\newblock


\bibitem[\protect\citeauthoryear{Chien, Peng, Li, and Milenkovic}{Chien
  et~al\mbox{.}}{2021}]%
        {GPR}
\bibfield{author}{\bibinfo{person}{Eli Chien}, \bibinfo{person}{Jianhao Peng},
  \bibinfo{person}{Pan Li}, {and} \bibinfo{person}{Olgica Milenkovic}.}
  \bibinfo{year}{2021}\natexlab{}.
\newblock \showarticletitle{Adaptive Universal Generalized PageRank Graph
  Neural Network}. In \bibinfo{booktitle}{\emph{International Conference on
  Learning Representations. https://openreview. net/forum}}.
\newblock


\bibitem[\protect\citeauthoryear{Coppersmith and Winograd}{Coppersmith and
  Winograd}{1987}]%
        {coppersmith1987matrix}
\bibfield{author}{\bibinfo{person}{Don Coppersmith} {and}
  \bibinfo{person}{Shmuel Winograd}.} \bibinfo{year}{1987}\natexlab{}.
\newblock \showarticletitle{Matrix multiplication via arithmetic progressions}.
  In \bibinfo{booktitle}{\emph{Proceedings of the nineteenth annual ACM
  symposium on Theory of computing}}. \bibinfo{pages}{1--6}.
\newblock


\bibitem[\protect\citeauthoryear{Deng, Dong, and Zhu}{Deng
  et~al\mbox{.}}{2019}]%
        {deng2019batch}
\bibfield{author}{\bibinfo{person}{Zhijie Deng}, \bibinfo{person}{Yinpeng
  Dong}, {and} \bibinfo{person}{Jun Zhu}.} \bibinfo{year}{2019}\natexlab{}.
\newblock \showarticletitle{Batch virtual adversarial training for graph
  convolutional networks}.
\newblock \bibinfo{journal}{\emph{arXiv preprint arXiv:1902.09192}}
  (\bibinfo{year}{2019}).
\newblock


\bibitem[\protect\citeauthoryear{Derr, Ma, Fan, Liu, Aggarwal, and Tang}{Derr
  et~al\mbox{.}}{2020}]%
        {EGCN}
\bibfield{author}{\bibinfo{person}{Tyler Derr}, \bibinfo{person}{Yao Ma},
  \bibinfo{person}{Wenqi Fan}, \bibinfo{person}{Xiaorui Liu},
  \bibinfo{person}{Charu Aggarwal}, {and} \bibinfo{person}{Jiliang Tang}.}
  \bibinfo{year}{2020}\natexlab{}.
\newblock \showarticletitle{Epidemic graph convolutional network}. In
  \bibinfo{booktitle}{\emph{Proceedings of the 13th International Conference on
  Web Search and Data Mining}}. \bibinfo{pages}{160--168}.
\newblock


\bibitem[\protect\citeauthoryear{Feng, Zhang, Dong, Han, Luan, Xu, Yang,
  Kharlamov, and Tang}{Feng et~al\mbox{.}}{2020}]%
        {GRAND}
\bibfield{author}{\bibinfo{person}{Wenzheng Feng}, \bibinfo{person}{Jie Zhang},
  \bibinfo{person}{Yuxiao Dong}, \bibinfo{person}{Yu Han},
  \bibinfo{person}{Huanbo Luan}, \bibinfo{person}{Qian Xu},
  \bibinfo{person}{Qiang Yang}, \bibinfo{person}{Evgeny Kharlamov}, {and}
  \bibinfo{person}{Jie Tang}.} \bibinfo{year}{2020}\natexlab{}.
\newblock \showarticletitle{Graph Random Neural Networks for Semi-Supervised
  Learning on Graphs}. In \bibinfo{booktitle}{\emph{Advances in Neural
  Information Processing Systems 33: Annual Conference on Neural Information
  Processing Systems 2020, NeurIPS 2020, December 6-12, 2020, virtual}}.
\newblock


\bibitem[\protect\citeauthoryear{Hamilton}{Hamilton}{2020}]%
        {hamilton2020graph}
\bibfield{author}{\bibinfo{person}{William~L Hamilton}.}
  \bibinfo{year}{2020}\natexlab{}.
\newblock \showarticletitle{Graph representation learning}.
\newblock \bibinfo{journal}{\emph{Synthesis Lectures on Artifical Intelligence
  and Machine Learning}} \bibinfo{volume}{14}, \bibinfo{number}{3}
  (\bibinfo{year}{2020}), \bibinfo{pages}{1--159}.
\newblock


\bibitem[\protect\citeauthoryear{Hamilton, Ying, and Leskovec}{Hamilton
  et~al\mbox{.}}{2017}]%
        {Graphsage}
\bibfield{author}{\bibinfo{person}{William~L. Hamilton},
  \bibinfo{person}{Zhitao Ying}, {and} \bibinfo{person}{Jure Leskovec}.}
  \bibinfo{year}{2017}\natexlab{}.
\newblock \showarticletitle{Inductive Representation Learning on Large Graphs}.
  In \bibinfo{booktitle}{\emph{Advances in Neural Information Processing
  Systems 30: Annual Conference on Neural Information Processing Systems 2017,
  December 4-9, 2017, Long Beach, CA, {USA}}}. \bibinfo{pages}{1024--1034}.
\newblock


\bibitem[\protect\citeauthoryear{Jin, Derr, Liu, Wang, Wang, Liu, and Tang}{Jin
  et~al\mbox{.}}{2020}]%
        {SelfTask}
\bibfield{author}{\bibinfo{person}{Wei Jin}, \bibinfo{person}{Tyler Derr},
  \bibinfo{person}{Haochen Liu}, \bibinfo{person}{Yiqi Wang},
  \bibinfo{person}{Suhang Wang}, \bibinfo{person}{Zitao Liu}, {and}
  \bibinfo{person}{Jiliang Tang}.} \bibinfo{year}{2020}\natexlab{}.
\newblock \showarticletitle{Self-supervised learning on graphs: Deep insights
  and new direction}.
\newblock \bibinfo{journal}{\emph{arXiv preprint arXiv:2006.10141}}
  (\bibinfo{year}{2020}).
\newblock


\bibitem[\protect\citeauthoryear{Kipf and Welling}{Kipf and Welling}{2017}]%
        {GCN}
\bibfield{author}{\bibinfo{person}{Thomas~N. Kipf} {and} \bibinfo{person}{Max
  Welling}.} \bibinfo{year}{2017}\natexlab{}.
\newblock \showarticletitle{Semi-Supervised Classification with Graph
  Convolutional Networks}. In \bibinfo{booktitle}{\emph{5th International
  Conference on Learning Representations, {ICLR}}}.
\newblock


\bibitem[\protect\citeauthoryear{Klicpera, Bojchevski, and
  G{\"{u}}nnemann}{Klicpera et~al\mbox{.}}{2019a}]%
        {APPNP}
\bibfield{author}{\bibinfo{person}{Johannes Klicpera},
  \bibinfo{person}{Aleksandar Bojchevski}, {and} \bibinfo{person}{Stephan
  G{\"{u}}nnemann}.} \bibinfo{year}{2019}\natexlab{a}.
\newblock \showarticletitle{Predict then Propagate: Graph Neural Networks meet
  Personalized PageRank}. In \bibinfo{booktitle}{\emph{7th International
  Conference on Learning Representations, {ICLR}}}.
\newblock


\bibitem[\protect\citeauthoryear{Klicpera, Wei{\ss}enberger, and
  G{\"{u}}nnemann}{Klicpera et~al\mbox{.}}{2019b}]%
        {Diffusion}
\bibfield{author}{\bibinfo{person}{Johannes Klicpera}, \bibinfo{person}{Stefan
  Wei{\ss}enberger}, {and} \bibinfo{person}{Stephan G{\"{u}}nnemann}.}
  \bibinfo{year}{2019}\natexlab{b}.
\newblock \showarticletitle{Diffusion Improves Graph Learning}. In
  \bibinfo{booktitle}{\emph{Advances in Neural Information Processing Systems
  32: Annual Conference on Neural Information Processing Systems 2019, NeurIPS
  2019, December 8-14, 2019, Vancouver, BC, Canada}}.
\newblock


\bibitem[\protect\citeauthoryear{Li, Han, and Wu}{Li et~al\mbox{.}}{2018a}]%
        {Oversmoothing}
\bibfield{author}{\bibinfo{person}{Qimai Li}, \bibinfo{person}{Zhichao Han},
  {and} \bibinfo{person}{Xiao{-}Ming Wu}.} \bibinfo{year}{2018}\natexlab{a}.
\newblock \showarticletitle{Deeper Insights Into Graph Convolutional Networks
  for Semi-Supervised Learning}. In \bibinfo{booktitle}{\emph{Proceedings of
  the Thirty-Second {AAAI} Conference on Artificial Intelligence}}.
\newblock


\bibitem[\protect\citeauthoryear{Li, Han, and Wu}{Li et~al\mbox{.}}{2018b}]%
        {li2018deeper}
\bibfield{author}{\bibinfo{person}{Qimai Li}, \bibinfo{person}{Zhichao Han},
  {and} \bibinfo{person}{Xiao-Ming Wu}.} \bibinfo{year}{2018}\natexlab{b}.
\newblock \showarticletitle{Deeper insights into graph convolutional networks
  for semi-supervised learning}. In \bibinfo{booktitle}{\emph{Proceedings of
  the AAAI Conference on Artificial Intelligence}}, Vol.~\bibinfo{volume}{32}.
\newblock


\bibitem[\protect\citeauthoryear{Liu, Gao, and Ji}{Liu et~al\mbox{.}}{2020}]%
        {DAGNN}
\bibfield{author}{\bibinfo{person}{Meng Liu}, \bibinfo{person}{Hongyang Gao},
  {and} \bibinfo{person}{Shuiwang Ji}.} \bibinfo{year}{2020}\natexlab{}.
\newblock \showarticletitle{Towards Deeper Graph Neural Networks}. In
  \bibinfo{booktitle}{\emph{{KDD} '20: The 26th {ACM} {SIGKDD} Conference on
  Knowledge Discovery and Data Mining}}.
\newblock


\bibitem[\protect\citeauthoryear{Milgram}{Milgram}{1967}]%
        {milgram1967small}
\bibfield{author}{\bibinfo{person}{Stanley Milgram}.}
  \bibinfo{year}{1967}\natexlab{}.
\newblock \showarticletitle{The small world problem}.
\newblock \bibinfo{journal}{\emph{Psychology today}} \bibinfo{volume}{2},
  \bibinfo{number}{1} (\bibinfo{year}{1967}), \bibinfo{pages}{60--67}.
\newblock


\bibitem[\protect\citeauthoryear{Murtagh}{Murtagh}{1991}]%
        {murtagh1991multilayer}
\bibfield{author}{\bibinfo{person}{Fionn Murtagh}.}
  \bibinfo{year}{1991}\natexlab{}.
\newblock \showarticletitle{Multilayer perceptrons for classification and
  regression}.
\newblock \bibinfo{journal}{\emph{Neurocomputing}} \bibinfo{volume}{2},
  \bibinfo{number}{5-6} (\bibinfo{year}{1991}), \bibinfo{pages}{183--197}.
\newblock


\bibitem[\protect\citeauthoryear{Page, Brin, Motwani, and Winograd}{Page
  et~al\mbox{.}}{1999}]%
        {page1999pagerank}
\bibfield{author}{\bibinfo{person}{Lawrence Page}, \bibinfo{person}{Sergey
  Brin}, \bibinfo{person}{Rajeev Motwani}, {and} \bibinfo{person}{Terry
  Winograd}.} \bibinfo{year}{1999}\natexlab{}.
\newblock \bibinfo{booktitle}{\emph{The PageRank citation ranking: Bringing
  order to the web.}}
\newblock \bibinfo{type}{{T}echnical {R}eport}. \bibinfo{institution}{Stanford
  InfoLab}.
\newblock


\bibitem[\protect\citeauthoryear{Pandit, Chau, Wang, and Faloutsos}{Pandit
  et~al\mbox{.}}{2007}]%
        {pandit2007netprobe}
\bibfield{author}{\bibinfo{person}{Shashank Pandit},
  \bibinfo{person}{Duen~Horng Chau}, \bibinfo{person}{Samuel Wang}, {and}
  \bibinfo{person}{Christos Faloutsos}.} \bibinfo{year}{2007}\natexlab{}.
\newblock \showarticletitle{Netprobe: a fast and scalable system for fraud
  detection in online auction networks}. In
  \bibinfo{booktitle}{\emph{Proceedings of the 16th international conference on
  World Wide Web}}. \bibinfo{pages}{201--210}.
\newblock


\bibitem[\protect\citeauthoryear{Paszke, Gross, Massa, Lerer, Bradbury, Chanan,
  Killeen, Lin, Gimelshein, Antiga, et~al\mbox{.}}{Paszke
  et~al\mbox{.}}{2019b}]%
        {paszke2019pytorch}
\bibfield{author}{\bibinfo{person}{Adam Paszke}, \bibinfo{person}{Sam Gross},
  \bibinfo{person}{Francisco Massa}, \bibinfo{person}{Adam Lerer},
  \bibinfo{person}{James Bradbury}, \bibinfo{person}{Gregory Chanan},
  \bibinfo{person}{Trevor Killeen}, \bibinfo{person}{Zeming Lin},
  \bibinfo{person}{Natalia Gimelshein}, \bibinfo{person}{Luca Antiga},
  {et~al\mbox{.}}} \bibinfo{year}{2019}\natexlab{b}.
\newblock \showarticletitle{Pytorch: An imperative style, high-performance deep
  learning library}.
\newblock \bibinfo{journal}{\emph{arXiv preprint arXiv:1912.01703}}
  (\bibinfo{year}{2019}).
\newblock


\bibitem[\protect\citeauthoryear{Paszke, Gross, Massa, Lerer, Bradbury, Chanan,
  Killeen, Lin, Gimelshein, Antiga, Desmaison, K{\"{o}}pf, Yang, DeVito,
  Raison, Tejani, Chilamkurthy, Steiner, Fang, Bai, and Chintala}{Paszke
  et~al\mbox{.}}{2019a}]%
        {pytorch}
\bibfield{author}{\bibinfo{person}{Adam Paszke}, \bibinfo{person}{Sam Gross},
  \bibinfo{person}{Francisco Massa}, \bibinfo{person}{Adam Lerer},
  \bibinfo{person}{James Bradbury}, \bibinfo{person}{Gregory Chanan},
  \bibinfo{person}{Trevor Killeen}, \bibinfo{person}{Zeming Lin},
  \bibinfo{person}{Natalia Gimelshein}, \bibinfo{person}{Luca Antiga},
  \bibinfo{person}{Alban Desmaison}, \bibinfo{person}{Andreas K{\"{o}}pf},
  \bibinfo{person}{Edward Yang}, \bibinfo{person}{Zachary DeVito},
  \bibinfo{person}{Martin Raison}, \bibinfo{person}{Alykhan Tejani},
  \bibinfo{person}{Sasank Chilamkurthy}, \bibinfo{person}{Benoit Steiner},
  \bibinfo{person}{Lu Fang}, \bibinfo{person}{Junjie Bai}, {and}
  \bibinfo{person}{Soumith Chintala}.} \bibinfo{year}{2019}\natexlab{a}.
\newblock \showarticletitle{PyTorch: An Imperative Style, High-Performance Deep
  Learning Library}. In \bibinfo{booktitle}{\emph{Advances in Neural
  Information Processing Systems 32: Annual Conference on Neural Information
  Processing Systems (NeurIPS) 2019}}. \bibinfo{pages}{8024--8035}.
\newblock


\bibitem[\protect\citeauthoryear{Pei, Wei, Chang, Lei, and Yang}{Pei
  et~al\mbox{.}}{2020}]%
        {geomgcn}
\bibfield{author}{\bibinfo{person}{Hongbin Pei}, \bibinfo{person}{Bingzhe Wei},
  \bibinfo{person}{Kevin~Chen{-}Chuan Chang}, \bibinfo{person}{Yu Lei}, {and}
  \bibinfo{person}{Bo Yang}.} \bibinfo{year}{2020}\natexlab{}.
\newblock \showarticletitle{Geom-GCN: Geometric Graph Convolutional Networks}.
  In \bibinfo{booktitle}{\emph{8th International Conference on Learning
  Representations, {ICLR} 2020, Addis Ababa, Ethiopia, April 26-30, 2020}}.
\newblock


\bibitem[\protect\citeauthoryear{Priebe, Shen, Huang, and Chen}{Priebe
  et~al\mbox{.}}{2020}]%
        {SGCC}
\bibfield{author}{\bibinfo{person}{Carey~E. Priebe}, \bibinfo{person}{Cencheng
  Shen}, \bibinfo{person}{Ningyuan Huang}, {and} \bibinfo{person}{Tianyi
  Chen}.} \bibinfo{year}{2020}\natexlab{}.
\newblock \showarticletitle{A Simple Spectral Failure Mode for Graph
  Convolutional Networks}.
\newblock \bibinfo{journal}{\emph{CoRR}}  \bibinfo{volume}{abs/2010.13152}
  (\bibinfo{year}{2020}).
\newblock
\showeprint[arxiv]{2010.13152}


\bibitem[\protect\citeauthoryear{Rong, Xu, Huang, Huang, Cheng, Ma, Wang, Derr,
  Wu, and Ma}{Rong et~al\mbox{.}}{2020}]%
        {rong2020deep}
\bibfield{author}{\bibinfo{person}{Yu Rong}, \bibinfo{person}{Tingyang Xu},
  \bibinfo{person}{Junzhou Huang}, \bibinfo{person}{Wenbing Huang},
  \bibinfo{person}{Hong Cheng}, \bibinfo{person}{Yao Ma}, \bibinfo{person}{Yiqi
  Wang}, \bibinfo{person}{Tyler Derr}, \bibinfo{person}{Lingfei Wu}, {and}
  \bibinfo{person}{Tengfei Ma}.} \bibinfo{year}{2020}\natexlab{}.
\newblock \showarticletitle{Deep graph learning: Foundations, advances and
  applications}. In \bibinfo{booktitle}{\emph{Proceedings of the 26th ACM
  SIGKDD International Conference on Knowledge Discovery \& Data Mining}}.
  \bibinfo{pages}{3555--3556}.
\newblock


\bibitem[\protect\citeauthoryear{Rossi, Jin, Kim, Ahmed, Koutra, and Lee}{Rossi
  et~al\mbox{.}}{2020}]%
        {proxmity}
\bibfield{author}{\bibinfo{person}{Ryan~A. Rossi}, \bibinfo{person}{Di Jin},
  \bibinfo{person}{Sungchul Kim}, \bibinfo{person}{Nesreen~K. Ahmed},
  \bibinfo{person}{Danai Koutra}, {and} \bibinfo{person}{John~Boaz Lee}.}
  \bibinfo{year}{2020}\natexlab{}.
\newblock \showarticletitle{On Proximity and Structural Role-based Embeddings
  in Networks: Misconceptions, Techniques, and Applications}.
\newblock \bibinfo{journal}{\emph{{ACM} Trans. Knowl. Discov. Data}}
  \bibinfo{volume}{14}, \bibinfo{number}{5} (\bibinfo{year}{2020}),
  \bibinfo{pages}{63:1--63:37}.
\newblock


\bibitem[\protect\citeauthoryear{Sen, Namata, Bilgic, Getoor, Galligher, and
  Eliassi-Rad}{Sen et~al\mbox{.}}{2008}]%
        {sen2008collective}
\bibfield{author}{\bibinfo{person}{Prithviraj Sen}, \bibinfo{person}{Galileo
  Namata}, \bibinfo{person}{Mustafa Bilgic}, \bibinfo{person}{Lise Getoor},
  \bibinfo{person}{Brian Galligher}, {and} \bibinfo{person}{Tina Eliassi-Rad}.}
  \bibinfo{year}{2008}\natexlab{}.
\newblock \showarticletitle{Collective classification in network data}.
\newblock \bibinfo{journal}{\emph{AI magazine}} \bibinfo{volume}{29},
  \bibinfo{number}{3} (\bibinfo{year}{2008}), \bibinfo{pages}{93--93}.
\newblock


\bibitem[\protect\citeauthoryear{Strassen}{Strassen}{1969}]%
        {strassen1969gaussian}
\bibfield{author}{\bibinfo{person}{Volker Strassen}.}
  \bibinfo{year}{1969}\natexlab{}.
\newblock \showarticletitle{Gaussian elimination is not optimal}.
\newblock \bibinfo{journal}{\emph{Numerische mathematik}} \bibinfo{volume}{13},
  \bibinfo{number}{4} (\bibinfo{year}{1969}), \bibinfo{pages}{354--356}.
\newblock


\bibitem[\protect\citeauthoryear{Velickovic, Fedus, Hamilton, Li{\`{o}},
  Bengio, and Hjelm}{Velickovic et~al\mbox{.}}{2019}]%
        {DGI}
\bibfield{author}{\bibinfo{person}{Petar Velickovic}, \bibinfo{person}{William
  Fedus}, \bibinfo{person}{William~L. Hamilton}, \bibinfo{person}{Pietro
  Li{\`{o}}}, \bibinfo{person}{Yoshua Bengio}, {and} \bibinfo{person}{R.~Devon
  Hjelm}.} \bibinfo{year}{2019}\natexlab{}.
\newblock \showarticletitle{Deep Graph Infomax}. In
  \bibinfo{booktitle}{\emph{7th International Conference on Learning
  Representations, {ICLR} 2019, New Orleans, LA, USA, May 6-9, 2019}}.
\newblock


\bibitem[\protect\citeauthoryear{Veličković, Cucurull, Casanova, Romero,
  Liò, and Bengio}{Veličković et~al\mbox{.}}{2018}]%
        {GAT}
\bibfield{author}{\bibinfo{person}{Petar Veličković},
  \bibinfo{person}{Guillem Cucurull}, \bibinfo{person}{Arantxa Casanova},
  \bibinfo{person}{Adriana Romero}, \bibinfo{person}{Pietro Liò}, {and}
  \bibinfo{person}{Yoshua Bengio}.} \bibinfo{year}{2018}\natexlab{}.
\newblock \showarticletitle{Graph Attention Networks}. In
  \bibinfo{booktitle}{\emph{International Conference on Learning
  Representations}}.
\newblock


\bibitem[\protect\citeauthoryear{Wang, Ying, Huang, and Leskovec}{Wang
  et~al\mbox{.}}{2020}]%
        {wang2020direct}
\bibfield{author}{\bibinfo{person}{Guangtao Wang}, \bibinfo{person}{Rex Ying},
  \bibinfo{person}{Jing Huang}, {and} \bibinfo{person}{Jure Leskovec}.}
  \bibinfo{year}{2020}\natexlab{}.
\newblock \showarticletitle{Direct Multi-hop Attention based Graph Neural
  Network}.
\newblock \bibinfo{journal}{\emph{arXiv preprint arXiv:2009.14332}}
  (\bibinfo{year}{2020}).
\newblock


\bibitem[\protect\citeauthoryear{Wang, Jin, and Derr}{Wang
  et~al\mbox{.}}{2021}]%
        {GNNBook-ch18-wang}
\bibfield{author}{\bibinfo{person}{Yu Wang}, \bibinfo{person}{Wei Jin}, {and}
  \bibinfo{person}{Tyler Derr}.} \bibinfo{year}{2021}\natexlab{}.
\newblock \showarticletitle{Graph Neural Networks: Self-supervised Learning}.
\newblock In \bibinfo{booktitle}{\emph{Graph Neural Networks: Foundations,
  Frontiers, and Applications}}, \bibfield{editor}{\bibinfo{person}{Lingfei
  Wu}, \bibinfo{person}{Peng Cui}, \bibinfo{person}{Jian Pei}, {and}
  \bibinfo{person}{Liang Zhao}} (Eds.). \bibinfo{publisher}{Springer},
  \bibinfo{address}{Singapore}, Chapter~18, \bibinfo{pages}{391--419}.
\newblock


\bibitem[\protect\citeauthoryear{Wu, Jr., Zhang, Fifty, Yu, and Weinberger}{Wu
  et~al\mbox{.}}{2019}]%
        {SGC}
\bibfield{author}{\bibinfo{person}{Felix Wu}, \bibinfo{person}{Amauri H.~Souza
  Jr.}, \bibinfo{person}{Tianyi Zhang}, \bibinfo{person}{Christopher Fifty},
  \bibinfo{person}{Tao Yu}, {and} \bibinfo{person}{Kilian~Q. Weinberger}.}
  \bibinfo{year}{2019}\natexlab{}.
\newblock \showarticletitle{Simplifying Graph Convolutional Networks}. In
  \bibinfo{booktitle}{\emph{Proceedings of the 36th International Conference on
  Machine Learning, {ICML} 2019, 9-15 June 2019, Long Beach, California,
  {USA}}}.
\newblock


\bibitem[\protect\citeauthoryear{Xu, Hu, Leskovec, and Jegelka}{Xu
  et~al\mbox{.}}{2019}]%
        {GIN}
\bibfield{author}{\bibinfo{person}{Keyulu Xu}, \bibinfo{person}{Weihua Hu},
  \bibinfo{person}{Jure Leskovec}, {and} \bibinfo{person}{Stefanie Jegelka}.}
  \bibinfo{year}{2019}\natexlab{}.
\newblock \showarticletitle{How Powerful are Graph Neural Networks?}. In
  \bibinfo{booktitle}{\emph{7th International Conference on Learning
  Representations, {ICLR} 2019, New Orleans, LA, USA, May 6-9, 2019}}.
\newblock


\bibitem[\protect\citeauthoryear{Xu, Li, Tian, Sonobe, Kawarabayashi, and
  Jegelka}{Xu et~al\mbox{.}}{2018}]%
        {JK}
\bibfield{author}{\bibinfo{person}{Keyulu Xu}, \bibinfo{person}{Chengtao Li},
  \bibinfo{person}{Yonglong Tian}, \bibinfo{person}{Tomohiro Sonobe},
  \bibinfo{person}{Ken{-}ichi Kawarabayashi}, {and} \bibinfo{person}{Stefanie
  Jegelka}.} \bibinfo{year}{2018}\natexlab{}.
\newblock \showarticletitle{Representation Learning on Graphs with Jumping
  Knowledge Networks}. In \bibinfo{booktitle}{\emph{Proceedings of the 35th
  International Conference on Machine Learning, {ICML} 2018,
  Stockholmsm{\"{a}}ssan, Stockholm, Sweden, July 10-15, 2018}}.
\newblock


\bibitem[\protect\citeauthoryear{Yan, Hashemi, Swersky, Yang, and Koutra}{Yan
  et~al\mbox{.}}{2021}]%
        {coin}
\bibfield{author}{\bibinfo{person}{Yujun Yan}, \bibinfo{person}{Milad Hashemi},
  \bibinfo{person}{Kevin Swersky}, \bibinfo{person}{Yaoqing Yang}, {and}
  \bibinfo{person}{Danai Koutra}.} \bibinfo{year}{2021}\natexlab{}.
\newblock \showarticletitle{Two Sides of the Same Coin: Heterophily and
  Oversmoothing in Graph Convolutional Neural Networks}.
\newblock \bibinfo{journal}{\emph{arXiv preprint arXiv:2102.06462}}
  (\bibinfo{year}{2021}).
\newblock


\bibitem[\protect\citeauthoryear{Yang, Cohen, and Salakhudinov}{Yang
  et~al\mbox{.}}{2016}]%
        {yang2016revisiting}
\bibfield{author}{\bibinfo{person}{Zhilin Yang}, \bibinfo{person}{William
  Cohen}, {and} \bibinfo{person}{Ruslan Salakhudinov}.}
  \bibinfo{year}{2016}\natexlab{}.
\newblock \showarticletitle{Revisiting semi-supervised learning with graph
  embeddings}. In \bibinfo{booktitle}{\emph{International conference on machine
  learning}}. PMLR, \bibinfo{pages}{40--48}.
\newblock


\bibitem[\protect\citeauthoryear{Ying, You, Morris, Ren, Hamilton, and
  Leskovec}{Ying et~al\mbox{.}}{2018}]%
        {Diffpool}
\bibfield{author}{\bibinfo{person}{Zhitao Ying}, \bibinfo{person}{Jiaxuan You},
  \bibinfo{person}{Christopher Morris}, \bibinfo{person}{Xiang Ren},
  \bibinfo{person}{William~L. Hamilton}, {and} \bibinfo{person}{Jure
  Leskovec}.} \bibinfo{year}{2018}\natexlab{}.
\newblock \showarticletitle{Hierarchical Graph Representation Learning with
  Differentiable Pooling}. In \bibinfo{booktitle}{\emph{Advances in Neural
  Information Processing Systems 31: Annual Conference on Neural Information
  Processing Systems 2018, NeurIPS 2018, December 3-8, 2018, Montr{\'{e}}al,
  Canada}}.
\newblock


\bibitem[\protect\citeauthoryear{Yuster and Zwick}{Yuster and Zwick}{2005}]%
        {yuster2005fast}
\bibfield{author}{\bibinfo{person}{Raphael Yuster} {and} \bibinfo{person}{Uri
  Zwick}.} \bibinfo{year}{2005}\natexlab{}.
\newblock \showarticletitle{Fast sparse matrix multiplication}.
\newblock \bibinfo{journal}{\emph{ACM Transactions On Algorithms (TALG)}}
  \bibinfo{volume}{1}, \bibinfo{number}{1} (\bibinfo{year}{2005}),
  \bibinfo{pages}{2--13}.
\newblock


\bibitem[\protect\citeauthoryear{Zeng, Zhang, Xia, Srivastava, Malevich,
  Kannan, Prasanna, Jin, and Chen}{Zeng et~al\mbox{.}}{2020}]%
        {zeng2020deep}
\bibfield{author}{\bibinfo{person}{Hanqing Zeng}, \bibinfo{person}{Muhan
  Zhang}, \bibinfo{person}{Yinglong Xia}, \bibinfo{person}{Ajitesh Srivastava},
  \bibinfo{person}{Andrey Malevich}, \bibinfo{person}{Rajgopal Kannan},
  \bibinfo{person}{Viktor Prasanna}, \bibinfo{person}{Long Jin}, {and}
  \bibinfo{person}{Ren Chen}.} \bibinfo{year}{2020}\natexlab{}.
\newblock \showarticletitle{Deep Graph Neural Networks with Shallow Subgraph
  Samplers}.
\newblock \bibinfo{journal}{\emph{arXiv preprint arXiv:2012.01380}}
  (\bibinfo{year}{2020}).
\newblock


\bibitem[\protect\citeauthoryear{Zhang and Chen}{Zhang and Chen}{2018}]%
        {SEAL}
\bibfield{author}{\bibinfo{person}{Muhan Zhang} {and} \bibinfo{person}{Yixin
  Chen}.} \bibinfo{year}{2018}\natexlab{}.
\newblock \showarticletitle{Link Prediction Based on Graph Neural Networks}. In
  \bibinfo{booktitle}{\emph{Advances in Neural Information Processing Systems
  31: Annual Conference on Neural Information Processing Systems (NeurIPS)
  2018}}.
\newblock


\bibitem[\protect\citeauthoryear{Zhou, Cui, Hu, Zhang, Yang, Liu, Wang, Li, and
  Sun}{Zhou et~al\mbox{.}}{2020}]%
        {zhou2020graph}
\bibfield{author}{\bibinfo{person}{Jie Zhou}, \bibinfo{person}{Ganqu Cui},
  \bibinfo{person}{Shengding Hu}, \bibinfo{person}{Zhengyan Zhang},
  \bibinfo{person}{Cheng Yang}, \bibinfo{person}{Zhiyuan Liu},
  \bibinfo{person}{Lifeng Wang}, \bibinfo{person}{Changcheng Li}, {and}
  \bibinfo{person}{Maosong Sun}.} \bibinfo{year}{2020}\natexlab{}.
\newblock \showarticletitle{Graph neural networks: A review of methods and
  applications}.
\newblock \bibinfo{journal}{\emph{AI Open}}  \bibinfo{volume}{1}
  (\bibinfo{year}{2020}), \bibinfo{pages}{57--81}.
\newblock


\bibitem[\protect\citeauthoryear{Zhu, Rossi, Rao, Mai, Lipka, Ahmed, and
  Koutra}{Zhu et~al\mbox{.}}{2020a}]%
        {zhu2020graph}
\bibfield{author}{\bibinfo{person}{Jiong Zhu}, \bibinfo{person}{Ryan~A Rossi},
  \bibinfo{person}{Anup Rao}, \bibinfo{person}{Tung Mai},
  \bibinfo{person}{Nedim Lipka}, \bibinfo{person}{Nesreen~K Ahmed}, {and}
  \bibinfo{person}{Danai Koutra}.} \bibinfo{year}{2020}\natexlab{a}.
\newblock \showarticletitle{Graph Neural Networks with Heterophily}.
\newblock \bibinfo{journal}{\emph{arXiv preprint arXiv:2009.13566}}
  (\bibinfo{year}{2020}).
\newblock


\bibitem[\protect\citeauthoryear{Zhu, Yan, Zhao, Heimann, Akoglu, and
  Koutra}{Zhu et~al\mbox{.}}{2020b}]%
        {zhu2020beyond}
\bibfield{author}{\bibinfo{person}{Jiong Zhu}, \bibinfo{person}{Yujun Yan},
  \bibinfo{person}{Lingxiao Zhao}, \bibinfo{person}{Mark Heimann},
  \bibinfo{person}{Leman Akoglu}, {and} \bibinfo{person}{Danai Koutra}.}
  \bibinfo{year}{2020}\natexlab{b}.
\newblock \showarticletitle{Beyond Homophily in Graph Neural Networks: Current
  Limitations and Effective Designs}.
\newblock \bibinfo{journal}{\emph{Advances in Neural Information Processing
  Systems}}  \bibinfo{volume}{33} (\bibinfo{year}{2020}).
\newblock


\end{thebibliography}










\end{document}